\documentclass{article}

\usepackage{arxiv}
\usepackage{wrapfig}

\usepackage[utf8]{inputenc} 
\usepackage[T1]{fontenc}    
\usepackage{hyperref}       
\usepackage{url}            
\usepackage{booktabs}       
\usepackage{amsfonts}       
\usepackage{nicefrac}       
\usepackage{microtype}      
\usepackage{dblfloatfix}
\usepackage{multirow}
\usepackage{lipsum}
\usepackage{graphicx}
\graphicspath{ {./images/} }
\usepackage{algorithm}%
\usepackage{algorithmic}%
\usepackage{subcaption}
\usepackage{amsmath}
\usepackage{xcolor}
\usepackage{array}    
\usepackage{tabularx} 
\usepackage{orcidlink}

\title{EasyDiagnos: A Framework for Accurate Feature Selection for Automatic Diagnosis in Smart Healthcare}

\author{
  \orcidlink{0000-0001-8057-6963} Prasenjit Maji \\
  Computer Science and Design\\
  Dr. B. C. Roy Engineering College\\India,WB, 713206 \\
 \texttt{maji.katm@gmail.com} \\
   \And
Amit Kumar Mondal \\
  Computer Science and Engineering\\
 Bengal College of Engg. and Tech.\\India,WB, 713212 \\
  \texttt{mamit5620@gmail.com} \\
  \And
 \orcidlink{0000-0002-9403-4724}Hemanta Kumar Mondal \\
  Electronics and Communication Engg.\\
  National Institute of Technology\\
  Durgapur,India, WB, 713209 \\
  \texttt{hkmondal.ece@nitdgp.ac.in} \\
   \And
 \orcidlink{0000-0003-2959-6541} Saraju P. Mahanty \\
  Dept. of Computer Science and Engineering\\
  University of North Texas, USA\\  
  \texttt{saraju.mohanty@unt.edu} \\
}

\begin{document}
\maketitle
\begin{abstract}
The rapid advancements in artificial intelligence (AI) have revolutionized smart healthcare, driving innovations in wearable technologies, continuous monitoring devices, and intelligent diagnostic systems. However, security, explainability, robustness, and performance optimization challenges remain critical barriers to widespread adoption in clinical environments. This research presents an innovative algorithmic method using the Adaptive Feature Evaluator (AFE) algorithm to improve feature selection in healthcare datasets and overcome problems.   AFE integrating Genetic Algorithms (GA), Explainable Artificial Intelligence (XAI), and Permutation and Combination Techniques (PCT), the algorithm optimizes Clinical Decision Support Systems (CDSS), thereby enhancing predictive accuracy and interpretability. The proposed method is validated across three diverse healthcare datasets using six distinct machine learning algorithms, demonstrating its robustness and superiority over conventional feature selection techniques. The results underscore the transformative potential of AFE in smart healthcare, enabling personalized and transparent patient care. Notably, the AFE algorithm, when combined with a Multi-layer Perceptron (MLP), achieved an accuracy of up to 98.5\%, highlighting its capability to improve clinical decision-making processes in real-world healthcare applications.
\end{abstract}


\section{Introduction}
The world has made medical illness analysis more and more critical \cite{higgins2020from}, leading to increased research and development efforts in this area. Advances in technology, such as deep learning and machine learning \cite{shortliffe2018clinical}, have enabled researchers to leverage the potential of healthcare data to create novel approaches that enhance human health outcomes. These algorithms can anticipate outcomes reasonably. However, those algorithms frequently need to explain their forecasts clearly, which reduces their effectiveness and reliability \cite{nazar2021systematic}. "Black Box" is the term or issue discussed about the reliability problem in \cite{von2021transparency}. Furthermore, these algorithms' lack of interpretability presents severe difficulties in clinical contexts where medical personnel need lucid explanations to comprehend and rely on the advice these models provide \cite{kuhn2017health}\cite{dwivedi2021artificial}. The urgent need for specific procedures in medical illness analysis is highlighted by this opacity, which not only makes adoption difficult but also raises questions about accountability and ethical issues  \cite{gerke2020ethical}\cite{reddy2020governance}. In response to this difficulty, Explainable AI (XAI) has surfaced to clarify the decision-making procedure of machine learning models, thereby offering discernment into the how and why of a given prediction \cite{angelov2020towards}. In medicine, XAI is essential for improving the dependability and comprehensibility of disease analysis models \cite{arrieta2020explainable}.

The process starts with selecting an extensive healthcare dataset to guarantee data quality and integrity. Next, detailed data analysis and preprocessing procedures are performed. The dataset is optimized for precise model training by researchers through data cleaning, normalization, and transformation.
After that, various deep learning and machine learning methods are use to create the model. These are comprised, however, and are not restricted to neural networks, decision trees, random forests, support vector machines, and ensemble techniques. By utilizing diverse algorithms, scholars can investigate every methodology's distinct advantages and drawbacks, consequently augmenting the probability of pinpointing the optimal model for illness forecasting.
Model explainability becomes the main concern when the models are trained. Several Explainable AI strategies are use to clarify the models' decision-making process. These methods include attention processes in deep learning models, SHAP (SHapley Additive exPlanations) values, feature importance analysis, and local interpretable model-agnostic explanations (LIME) \cite{hrnjica2020explainable}.

Inspired by natural selection, genetic algorithms are widely used to iteratively identify optimal feature subsets that improve machine learning model performance. Likewise, permutation and combination techniques provide a thorough method for feature selection by evaluating all possible feature subsets using a fitness function. 
Using these techniques, researchers can obtain essential insights into the fundamental principles underlying the predictions, improving comprehension and confidence in the model's results.

Integrating XAI, GA, and PCT enhances model interpretation and refinement by identifying biases, outliers, and confounding factors, thereby improving the robustness of disease analysis models. This iterative process enables researchers to select the most suitable model for real-world healthcare applications, driving innovation toward more reliable, transparent, and interpretable healthcare analytics systems. The proposed Adaptive Feature Evaluator (AFE) is a novel algorithm designed to assess feature significance across various datasets. Through rigorous testing on diverse datasets using both machine learning and deep learning algorithms, the AFE has consistently outperformed existing feature selection methods. One of the critical strengths of AFE is its adaptability, making it compatible with any algorithm and dataset. Our approach integrates three popular and highly effective feature selection strategies: XAI, PCT, and GA. By combining these techniques, the AFE computes accuracy scores for each method individually, consistently demonstrating superior performance compared to traditional feature selection algorithms.

The salient contributions of the proposed work are as follows: 
\begin{itemize}
    \item Development of the Adaptive Feature Evaluator (AFE): The paper introduces a novel AFE algorithm that integrates Genetic Algorithms (GA), Permutation \& Combination Techniques (PCT), and Explainable AI (XAI) to improve feature selection, specifically enhancing Clinical Decision Support Systems (CDSS).
    \item Superiority in Performance: AFE consistently outperforms traditional feature selection methods, achieving up to 98.5\% accuracy across diverse datasets and algorithms, demonstrating its robustness and effectiveness.
    \item Versatility and Adaptability: The AFE's adaptability allows it to be effectively applied across various algorithms and datasets, broadening its utility in healthcare analytics.
    \item Impact on Healthcare: The work advances healthcare analytics by providing more reliable, transparent, and interpretable systems for clinical decision-making, contributing to better patient care.
\end{itemize}
Overall, the contribution of this work lies in developing the AFE algorithm, which combines cutting-edge feature selection techniques with XAI to enhance the accuracy, transparency, and robustness of clinical decision support systems, thereby advancing the field of healthcare analytics.

The remainder of this paper is organized as follows: Section 2 outlines the unique contributions of our research and Section 3 reviews related literature. Section 4  explores the challenges faced in implementing Explainable AI (XAI). Section 5 provides the foundational background necessary for understanding our work along with the proposed framework. Section 6 details the experimental procedures and presents the results. Lastly, Section 7 offers concluding remarks on our study.

\section {Contribution of the Work}
\label{sec:contribution}
\subsection{Problems Addressed in the Current Work}
\begin{figure*}[!t]
\centering
\includegraphics[width=6.0in]{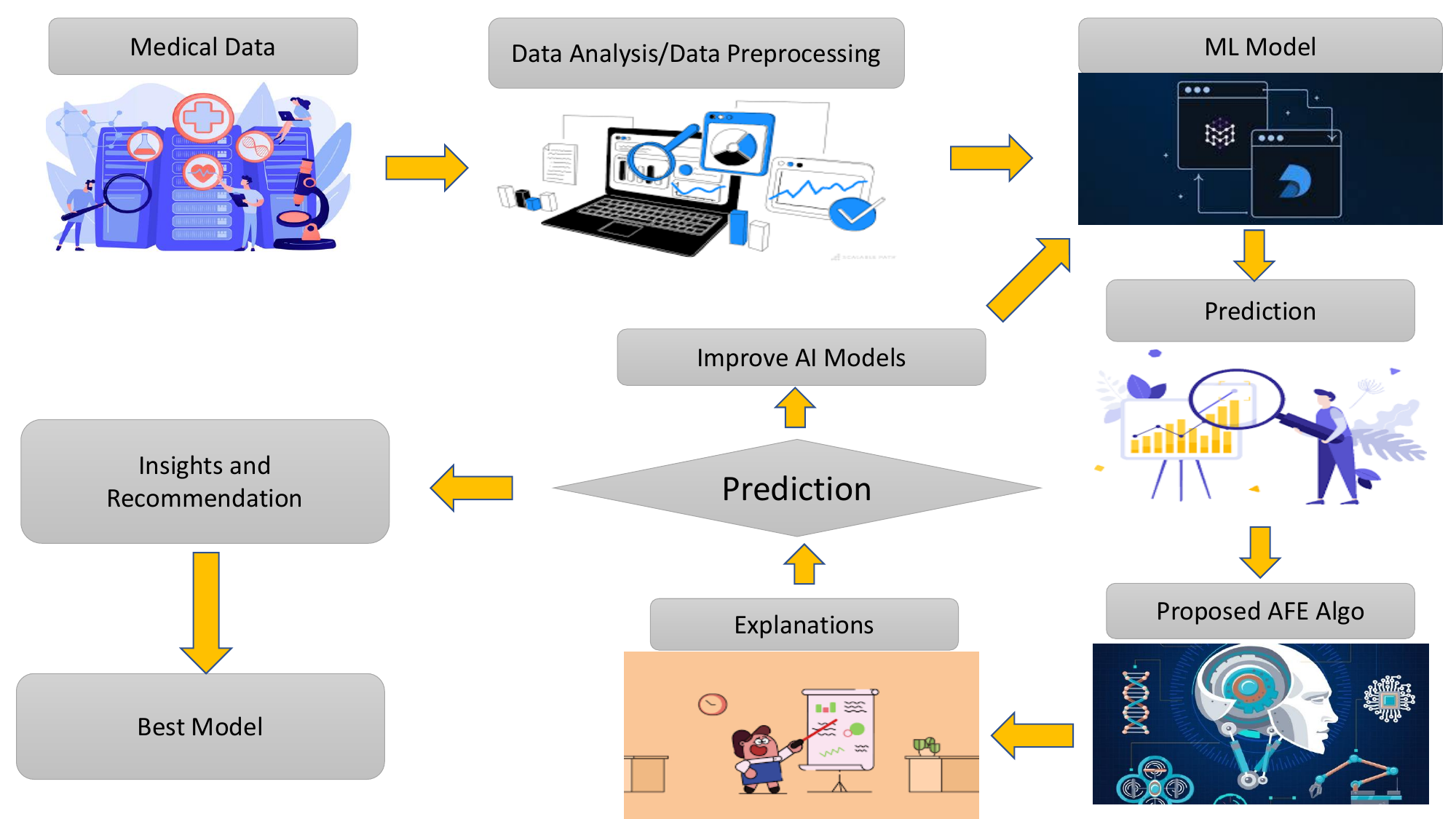}
\caption{System Model for Integrating Explainable AI in Clinical Decision Support Systems}
\label{fig1}
\end{figure*}

A more thorough investigation and comparison of various algorithms must be conducted. Many existing studies rely on one or two algorithms without validating their effectiveness comprehensively. Such implementations raise questions about whether these algorithms are the most suitable for the task. The current work aims to fill these gaps by introducing the Adaptive Feature Evaluator (AFE), which integrates multiple effective feature selection techniques and enhances model interpretability, providing a more robust and transparent approach to medical data analysis. Fig. \ref{fig1} provides a concise summary of our study. The initial phase in the proposed paradigm is the gathering and preparation of medical data. After processing, a machine learning (ML) model is trained using this data to produce predictions. By integrating XAI, GA, and PCT, valuable insights and recommendations can be generated. The AFE algorithm is then applied to select the most appropriate approach for clinical decision support.  
\subsection{Solution Proposed}
The proposed study addresses the explainability issue by integrating GA, XAI, and PCT. This approach aims to enhance understanding of the prediction processes by leveraging these methodologies to provide insights into the inner workings of deep learning and machine learning algorithms. By combining, the study improves model transparency and simplifies modifications, impacting the healthcare sector significantly.

Additionally, the study tackles the problem of limited algorithm exploration by introducing a novel algorithm that leverages feature-based techniques. This new method utilizes various feature set algorithms to generate predictions for the same objectives. It identifies the most suitable model for a given dataset through hyperparameter tuning and comprehensive analysis, ensuring a thorough evaluation of algorithmic performance and enhancing healthcare applications' accuracy and predictive efficacy.
\subsection{Novelty of the Work}
The paper introduces the Adaptive Feature Evaluator (AFE), a novel algorithm that integrates Genetic Algorithms, Permutation \& Combination Techniques, and Explainable AI  to enhance feature selection and improve Clinical Decision Support Systems. The AFE achieves up to 98.5\% accuracy across diverse datasets and algorithms, demonstrating its robustness and effectiveness. Its adaptability allows for broad application across various algorithms and datasets, expanding its utility in healthcare analytics.

Additionally, the paper presents a novel approach to predictive modeling by employing six distinct machine learning methods and neural networks rather than limiting the investigation to a few algorithms. This diverse methodological approach provides multiple perspectives on the same dataset, offering subtle insights that can enhance the efficacy of treatment plans and diagnostics. This strategy advances healthcare analytics and opens avenues for improved predictive modeling through algorithmic diversity, contributing to better patient care.
\section {Related Prior Research}
The increasing demand for personalized healthcare solutions has driven significant advancements in smart healthcare devices, offering continuous and automatic monitoring capabilities. Recent developments, such as MyWear, a novel smart garment that enables continuous vital monitoring, demonstrate the potential of wearable technology in healthcare, providing real-time health data with minimal patient intervention \cite{sethuraman2021mywear}. The study employed post-hoc and agnostic models, namely Local Interpretable Model-Agnostic Explanations (LIME) and SHapley Additive exPlanations (SHAP), to determine the most significant genes for classifying lung cancer types and subtypes \cite{ramos2021interpretable}, as well as the most crucial features for predicting lung cancer survival \cite{siddhartha2019explanatory}. Intelligent devices like iKardo, an advanced ECG monitoring system, enhance competent healthcare by automatically identifying critical heartbeats, thereby aiding in timely interventions for cardiovascular conditions \cite{maji2021ikardo}. Additionally, continuous glucose monitoring technologies have evolved to provide patients and healthcare providers with comprehensive and actionable insights into glucose levels, supporting diabetes management with a high degree of accuracy and convenience \cite{joshi2021cgm}. These innovations underscore the importance of integrating advanced algorithms and smart devices in developing personalized and efficient healthcare solutions.

In recent times, the researchers discussed, for instance, and suggested utilizing SHAP and LIME in conjunction with iAFPs-EnC-GA (GA (Fuzzy K-nearest neighbor (FKNN), Random Forest (RF), k-nearest neighbor (KNN), and Support Vector Machine (SVM)) for fungal infection \cite{ahmad2022iafps}. The paper introduces a deep ensemble method that uses uncertainty in relevance scores to improve the reliability and trustworthiness of predictions for clinical time series data with explainable AI \cite{wickstrom2021uncertainty}. 

Mediastinal Cysts and Tumors have also been detected using the Ensemble of Extreme Gradient Boosting (XGBoost) and SHAP \cite{wang2021radiomics}. Recently, bloodstream infections with SHAP(XAI) have been found in \cite{pai2021artificial} using XGBoost, RF, SVM, and MLP.
An SVM-based model for predicting lung cancer from an image dataset is proposed in \cite{kumar2022lung}. Using a dataset from the University of California's online repository, they performed preprocessing operations, such as eliminating values that are not relevant. The work separated the dataset into training and testing sets and used the SVM model to forecast using the features retrieved. With an accuracy of almost 98.8\%, the model outperformed KNN, Naive Bayes, and J48 models. The study introduces a scalable ML model using minimal cognitive tests for accurate and explainable dementia risk prediction in aging populations. \cite{wang2021efficient}.

The study employs DNN and XAI (SHAP) to predict postprandial glucose levels in Type 1 diabetes, enhancing artificial pancreas systems and decision-making tools. \cite{annuzzi2024exploring}.
The researchers propose a histopathology image gathered from the LC25000 dataset to construct a CAD system for lung and colon cancer analysis  \cite{chehade2022lung}. Hu invariant moments and the GLCM method extract features from the pictures. The authors have classified lung and colon cancer using MLP, XGBoost, RF, SVM, and LDA models. Furthermore, the expected results of the ML models are explained using the SHAP technique. With 98.8\% and 99\% accuracy, respectively, XGBoost has the highest F1 score and accuracy out of the five models.
The author on the domain in recent times \cite{ismail2021lung} suggested a technique for early-stage lung cancer diagnosis that uses three picture datasets and machine learning models. Using CNN for classification and U-Net CNN for lung nodule segmentation, they obtained an AUC of 0.6459. 

Generative AI has been explored as a means to enhance data quality, demonstrating its effectiveness as a complementary tool to traditional methods. It has been shown to improve accuracy and streamline workflows. Integrating Generative AI into data quality processes can yield substantial long-term organizational benefits, including increased efficiency and enhanced decision-making capabilities \cite{Dhoni2023}. Feature selection is essential for removing irrelevant features, enhancing model accuracy, and reducing costs, as reviewed in key literature \cite{Thomas2020}. The Integrate-RF approach, a hybrid feature selection method combining multiple models with random forests, addresses dimensionality reduction and overfitting by selecting optimal features based on Out-of-Bag (OOB) classification error rates \cite{Wang2021}. A novel neural network-based feature selection method employing a weighting approach is proposed to highlight critical features, significantly enhancing algorithm speed and accuracy \cite{Challita2016}. Another study presents a correction method for feature importance bias in RandomForest models using permutation-based p-values, which improves interpretability and prediction accuracy by identifying significant variables in simulated and real-world datasets \cite{altmann2010permutation}. Using data from 5,601 COVID-19 patients in South Korea, an AI model is developed to predict clinical severity levels—categorized as low or high—based on 20 critical features identified through feature importance analysis. The model, constructed with a 5-layer deep neural network, was evaluated using metrics such as accuracy, specificity, and AUC \cite{Chung2021}. A two-stage surrogate-assisted evolutionary approach is introduced to reduce computational costs in Genetic Algorithm (GA)--based feature selection for large datasets. The approach utilizes a qualitative meta-model, for instance, selection, with CHC-QX and PSO-QX algorithms achieving faster convergence and greater accuracy, especially with datasets exceeding 100K instances \cite{Altarabichi2023}.
\begin{table}[htbp]
\centering
\caption{Review of the literature in reference to the proposed study}
\label{Tab1}
\begin{tabular}{cccc}
\hline
\hline
Work  & Dataset Name & Algorithm Used & Accuracy  \\
\hline
Y.Li et al.(2020)\cite{li2020classify} & LUNA16 Lung  & CNN with  & 82.15\% \\
&Cancer Dataset&Resnet-18&\\[1ex]
Ansari et al.(2011)\cite{ansari2011predictive} & UCI Heart & Bayes Net  & 87\% \\
&disease dataset&&\\[1ex]
Beyene at al.(2018)\cite{beyene2018survey} & UCI Heart  & ANN and SVM & 81.82\% and  \\&disease dataset&&80.38\%\\[1ex]
Riaz at al.(2018)\cite{riaz2018prediction} & UCI Heart  & KNN and Decision  & 78\% and 80\% \\
&disease dataset&Tree&\\[1ex]
Gularia et al.(2022)\cite{guleria2022xai} & Cardiovascular  & LR, SVM and  & 81.2\%, 82.5\%  \\
&Disease Dataset&KNN with XAI&and  75.9\%\\[1ex]
Patro et al.(2018)\cite{patro2023secure} & Heart Disease  & RF with  & 87\% \\
&Test & of XAI(SHAP and LIME)&\\[1ex]
Batista et al.(202)\cite{batista2020covid} & Covid-19 pandemic  & LR, RF  & Near about \\
&Dataset&and SVM&85\%\\[1ex]
Mahdy et al.(2020)\cite{mahdy2020automatic} & Covid-19  & Multi-level Thresholding   & 95\% \\
&pandemic dataset&with SVM&\\[1ex]
Ahmad et al.(2023)\cite{ahmed2023interpretable} & Lung Cancer  & DT, LR,   & 95\%, 97\%,  \\&Dataset&RF and Naïve Bayes &97\% and 92\%\\&&with XAI(SHAP \& LIME)&\\[1ex]
Malafaia et al.(2021)\cite{malafaia2021robustness} & LIDC-IDRI   & CNN with help & 89.60\% \\
&Lung Cancer& of XAI&\\[1ex]
\hline
\hline
\end{tabular}
\end{table}
\section{Challenges in Deploying XAI for Healthcare Applications}
The development of AI systems that can provide clear, understandable justifications for their choices and actions is known as Explainable Artificial Intelligence (XAI). Incorporating XAI in the healthcare setting presents many obstacles \cite{allen2023discovering}\cite{giuste2023explainable}.

Data related to healthcare is, by its very nature, complex, multifaceted, and derived from multiple sources. It is tough to integrate and interpret the data in a way that makes sense to build transparent AI models \cite{ali2023metaverse}.

Because of their complex structures, advanced artificial intelligence models, such as those based on deep learning, frequently serve as "black boxes." One major challenge is explaining these models' decision-making procedures in a way that medical professionals can comprehend \cite{ding2021boosting}.

A key component of interpretability is recognizing and displaying the most important factors affecting a choice. Understanding which patient characteristics the AI model considers and how it affect predictions is critical in the healthcare industry \cite{linardatos2021explainable}.

The healthcare industry operates in a highly regulated space, XAI systems must abide by strict laws like the Health Insurance Portability and Accountability Act. Applying models in many domains becomes more challenging when achieving regulatory criteria while ensuring model correctness and interpretability \cite{rane2023printing}.
It might be easier to create interfaces that efficiently tell healthcare professionals about AI-generated insights by overloading them with data. Achieving effective human-AI engagement requires finding the ideal balance between offering comprehensive explanations and preserving simplicity \cite{zhang2022explainable}.
\section {The Proposed Novel EasyDiagnos Framework}
\begin{figure*}[ht]
\centering
\includegraphics[width=6.0in]{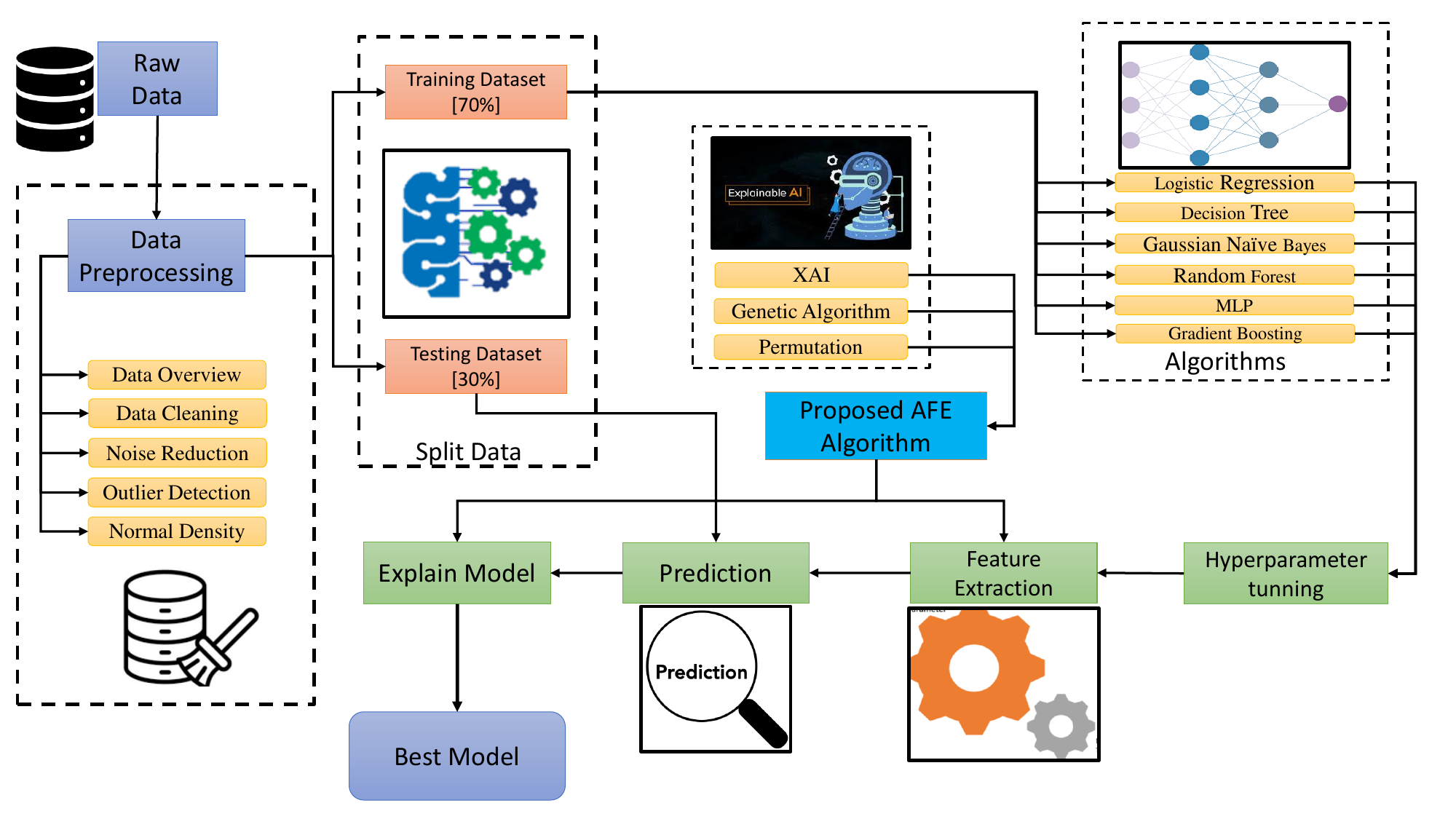}
\caption{Workflow framework of the proposed approach}
\label{fig2}
\end{figure*}

This section outlines the foundational components of our study, including the architecture of XAI, various Machine Learning (ML) frameworks, and the EasyDiagnos framework. Fig. \ref{fig2} illustrates the workflow of the proposed EasyDiagnos framework, detailing the entire implementation process. It begins with data augmentation, progresses through ML implementation and feature importance evaluation using the proposed Adaptive Feature Evaluator (AFE) algorithm, and concludes with the selection of the optimal model.

\subsection{What is XAI?}
Explainable AI (XAI) is the term used to describe a group of artificial intelligence systems that have the ability to explain their own activities, revealing their advantages, disadvantages, and possible future behaviors. XAI's main philosophy is to use a variety of approaches. These methods are meant to provide future developers with a wide range of design choices that compromise explainability and performance. Essentially, XAI aims to improve artificial intelligence system transparency by providing insights into algorithms' decision-making processes, making it easier to understand their outputs, and building user and stakeholder trust.

\subsection{Genetic Algorithm and PCT for Feature Selection}
A genetic algorithm for feature selection is a search heuristic that uses principles of natural selection and genetics to find the optimal subset of features that improves the performance of a machine learning model. Permutation and combination algorithms are exhaustive search methods used for feature selection. They involve generating all possible subsets of features and evaluating their performance using a fitness function. Algorithm \ref{alg:GA} represents the working process of the GA process, and Algorithm \ref{alg:permutation_combination} provides the approach for the PCT.

\begin{algorithm}
\caption{Adaptive Feature Selection Using Genetic Algorithm (GA)}
\label{alg:GA}

\begin{algorithmic}[1]
    \STATE \textbf{Input:} Dataset $D$ with features $\{A, B, C, D, E\}$ and target $T$
    \STATE \textbf{Output:} Best subset of features $S_{best}$
    \STATE \textbf{Initialize:} Set population size $N$, top solutions $k$, crossover rate $P_c$, mutation rate $P_m$, maximum iterations $MaxIter$
    
    \STATE Generate initial population $P = \{S_1, S_2, \ldots, S_N\}$ where $S_i \subseteq \{A, B, C, D, E\}$
    
    \STATE \textbf{Evaluate Fitness:} For each $S_i \in P$, compute the fitness (e.g., accuracy)
    
    \STATE \textbf{Selection:} Sort population $P$ by fitness, select the top $k$ subsets to form $P_{selected}$
    
    \STATE \textbf{Crossover:} For each pair $(S_i, S_j) \in P_{selected}$, perform crossover with probability $P_c$ to generate new subsets $S_{new}$
    
    \STATE \textbf{Mutation:} For each $S_{new}$, mutate by adding or removing a feature with probability $P_m$
    
    \STATE \textbf{Replacement:} Evaluate fitness of $S_{new}$ and replace the least fit subsets in $P$ with the new ones
    
    \STATE \textbf{Termination:} If $MaxIter$ or a satisfactory fitness level is reached, stop; otherwise, repeat
    
    \STATE \textbf{Output:} Return the subset $S_{best}$ with the highest fitness from the final population
\end{algorithmic}
\end{algorithm}


\subsection{Exploration of Classification Techniques}
\subsubsection{Logistic Regression}
A statistical technique for examining a dataset where one or more independent factors influence a result is called logistic regression. It does this by fitting data to a logistic curve, which estimates the likelihood that an event will occur.
The logistic function is given by:
\begin{equation}
P(y=1|x) = \frac{1}{1 + e^{-z}}
\end{equation}
where
\begin{equation}
z = w^Tx + b
\end{equation}
\( w \) are the weights, \( x \) is the input features, and \( b \) is the bias term.

\subsubsection{Decision Tree}
Models such as decision trees are utilized for both regression and classification. The dataset is divided into smaller subsets according to the most important attributes, resulting in a structure resembling a tree.

\subsubsection{Gaussian Naive Bayes}
Naive Bayes is a probabilistic classifier that relies on the premise of feature independence and is based on the Bayes theorem.
The mathematical expression is given by:
\begin{equation}
P(y|x) = \frac{P(x|y)P(y)}{P(x)}
\end{equation}
where \( P(y|x) \) is the posterior probability, \( P(x|y) \) is the likelihood, \( P(y) \) is the prior probability, and \( P(x) \) is the evidence.

\subsubsection{Random Forest} 
During training, the Random Forest ensemble learning technique creates a large number of decision trees and outputs the class mode. Random Forest combines multiple decision trees, each built using a random subset of features and data samples.

\subsubsection{Multi-layer Perceptron (MLP) classifier} 
Multiple layers of nodes make up the MLP kind of feedforward neural network, which may learn non-linear correlations.
The mathematical expression for hidden layer activations is given by:
\begin{equation}
h_i = \sigma\left(\sum_{j=1}^{m} w_{ij}^{(1)}x_j + b_i^{(1)}\right)
\end{equation}
where \( \sigma \) is the activation function, \( w_{ij}^{(1)} \) are the weights, \( x_j \) are the inputs, and \( b_i^{(1)} \) are the biases.

\subsubsection{Gradient Boosting} 
Gradient Boosting is an ensemble strategy in which errors caused by previously trained models are corrected by adding weak learners one after the other.
The mathematical expression is given by:
\begin{equation}
F_m(x) = F_{m-1}(x) + \arg \min_{h} \sum_{i=1}^{N} L(y_i, F_{m-1}(x_i) + h(x_i))
\end{equation}
where \( F_m(x) \) is the model at iteration \( m \), \( h \) is the weak learner, and \( L \) is the loss function.

\begin{algorithm}
\caption{Feature Selection Using Permutation and Combination Technique}
\label{alg:permutation_combination}
\begin{algorithmic}[1]
    \STATE \textbf{Input:} Dataset with features $\{A, B, C\}$ and target variable
    \STATE \textbf{Output:} Best subset of features for predicting the target variable
    \STATE \textbf{Step 1: Permutation Generation:} Generate all possible permutations of features $\{A, B, C\}$
    \STATE Permutations: $\{A, B, C\}$, $\{A, C, B\}$, $\{B, A, C\}$, $\{B, C, A\}$, $\{C, A, B\}$, $\{C, B, A\}$
    \STATE \textbf{Step 2: Combination Generation:} Generate all possible combinations of features $\{A, B, C\}$
    \STATE Combinations: $\{A\}$, $\{B\}$, $\{C\}$, $\{A, B\}$, $\{A, C\}$, $\{B, C\}$, $\{A, B, C\}$
    \STATE \textbf{Step 3: Fitness Evaluation:} Evaluate the fitness of each permutation and combination using a fitness function (e.g., accuracy)
    \STATE Compute fitness for each subset
    \STATE \textbf{Step 4: Selection:} Select the subset with the highest fitness score as the best subset
    \STATE \textbf{Output:} The best subset of features for predicting the target variable
\end{algorithmic}
\end{algorithm}
\subsection{Description of Interpretable Models}
Contextual Importance and Utility (CIU), Gradient-Weighted Class Activation Mapping (Grad-CAM), SHapley Additive exPlanations (SHAP), and Local Interpretable Model-agnostic Explanations (LIME) are key approaches in XAI that enhance understanding DL and ML algorithms. Among these, SHAP and LIME are particularly recognized for their ability to improve model accuracy and clarify model behavior. In this work, we employ both SHAP and LIME as the primary part of the AFE. By leveraging these advanced XAI techniques, our objective is to deliver clear and interpretable model outputs, thereby fostering stakeholder confidence, enabling informed decision-making, and bridging the gap between complex AI models and human understanding.
\subsubsection{SHapley Additive exPlanations (SHAP)}
For individuals working on machine learning models or AI interpretation, SHAP (Shapley Additive Explanations) is essential for understanding model predictions. Because SHAP is flexible and can explain predictions from any model, it is a well-liked option in XAI libraries. SHAP provides a thorough approach to interpretation, drawing on contributions from other XAI approaches like LIME, SHAPely, Sampling Values, and DeepLift. Fundamentally, SHAP works by calculating Shapley values for every feature in the dataset used to train and evaluate the machine learning model being examined. These Shapley values provide essential insights into the model's decision-making process by quantifying the impact of each feature on the predictions it generates.

In a simplified form, the SHAP value for a feature \( i \) in a prediction \( x \) can be expressed as equation 7.
\begin{equation}
    \phi_i(x) = \sum_{S \subseteq \{1,2,...,p\} \setminus \{i\}} \frac{|S|!(p-|S|-1)!}{p!} [f_x(S \cup \{i\}) - f_x(S)]
\end{equation}

Here, 
 \( \phi_i(x) \) represents the SHAP value for feature \( i \) in prediction \( x \)
and \( f_x(S) \) represents the model's output when considering only the features in subset \( S \).
 \( p \) represents the total number of features.
 
 The SHAP library offers a variety of explainers, each designed to fit particular model types and data properties. A suitable explainer must be chosen for AI models and datasets to be effectively interpreted. Among these explainer kinds are: i)shape.Explainer ii)Treeshape. explainers. iii)Shape.Elucidators.Linear iv)Explainers of shape.Alternation v)Shape.Elucidators.Deep
\begin{figure}[!t]
\centering
\includegraphics[width=4.4in]{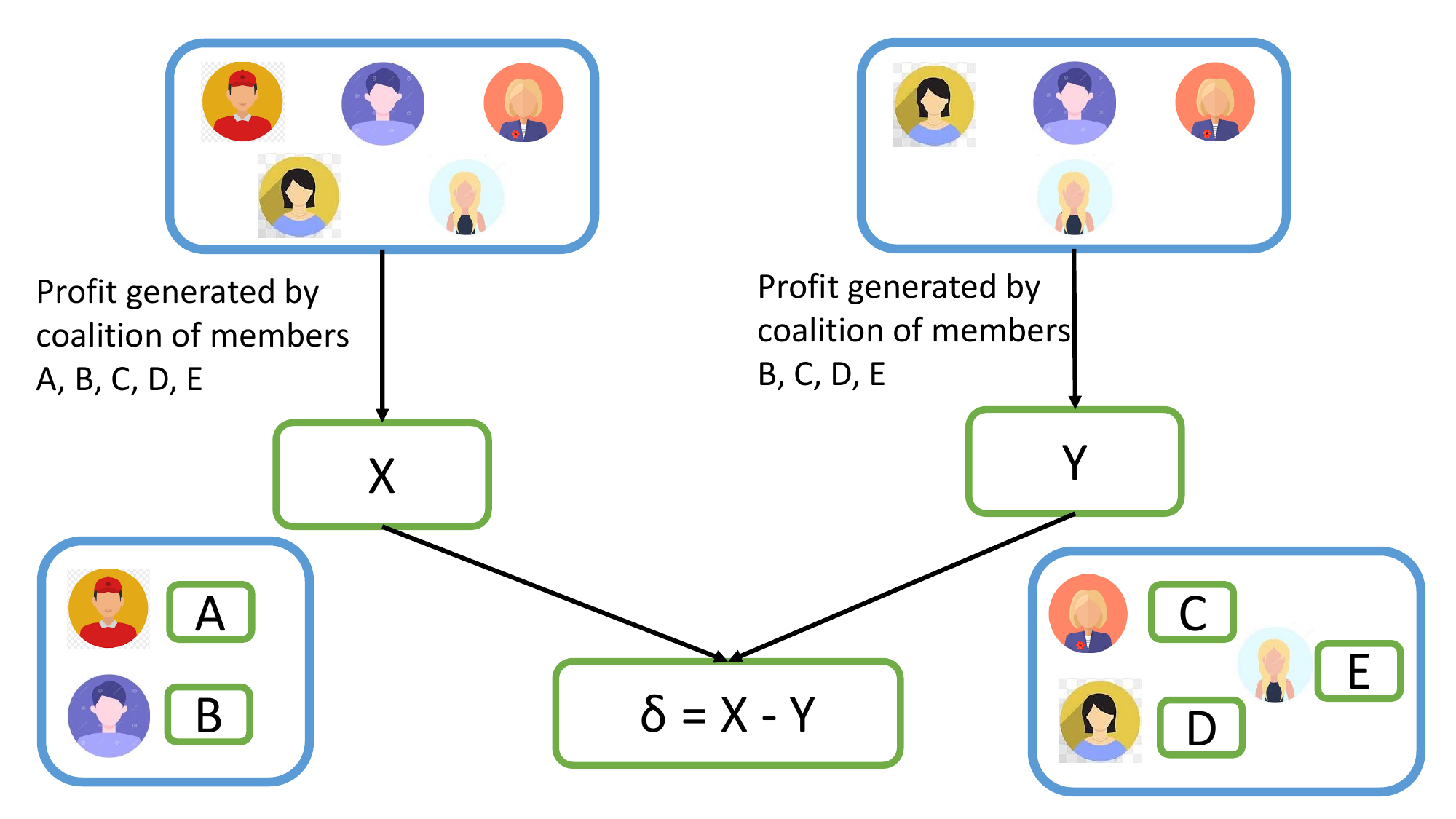}
\caption{Calculating the marginal distribution for a feature}
\label{figS1}
\end{figure}
\begin{figure}[!t]
\centering
\includegraphics[width=4.4in]{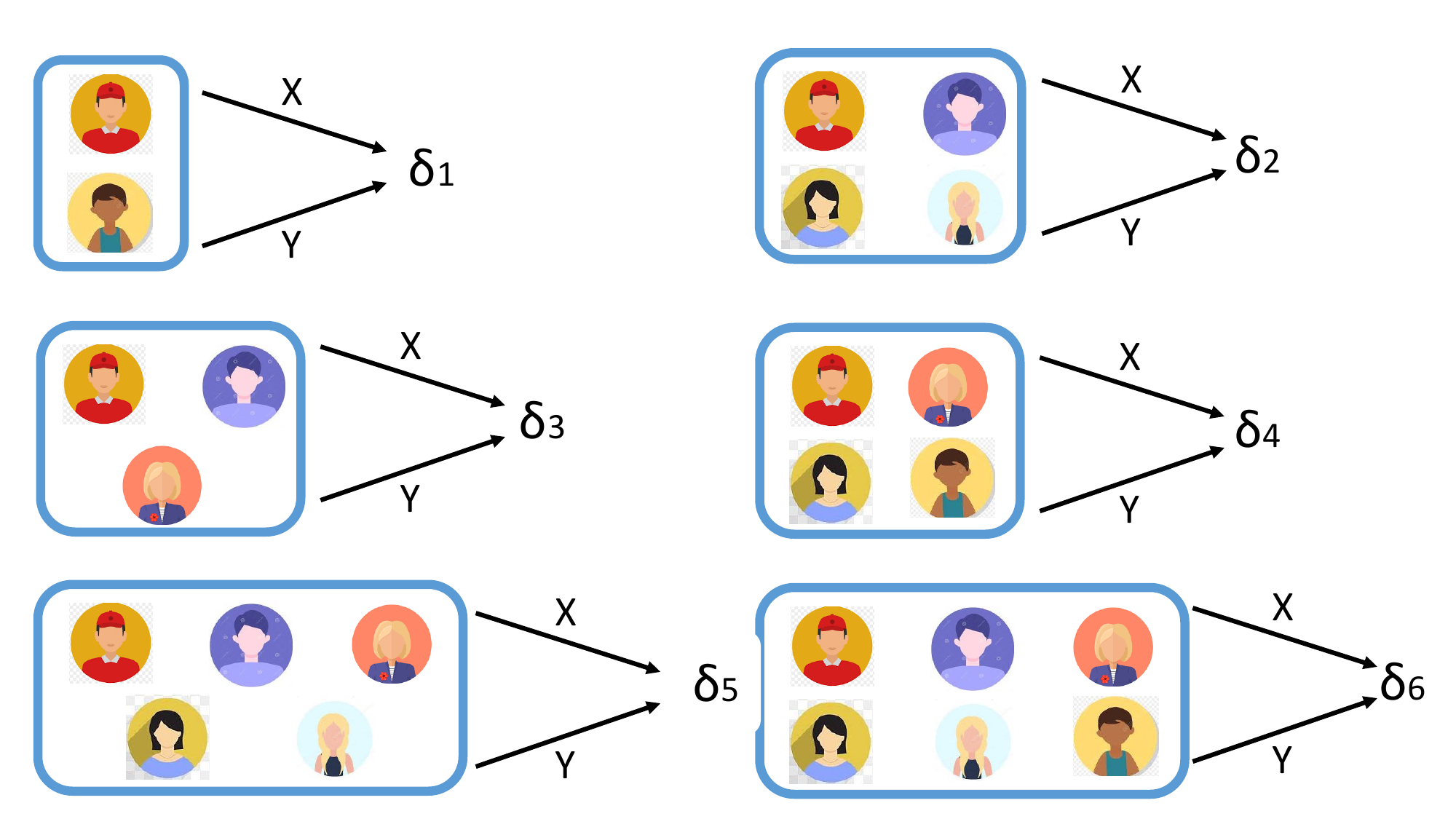}
\caption{Calculating the shapley value for a feature}
\label{figS2}
\end{figure}
\subsubsection{SHAP Value Explain By An Example} 
Shapley's values are rooted in cooperative game theory, which assesses each player's contribution to a game. This idea involves allocating prizes to 'n' participants in an equitable manner according to each player's unique contribution, much to Shapley Value.
The average marginal contribution of a characteristic for a particular instance, across all conceivable combinations inside the sample, is calculated by the Shapley value. This method guarantees a sophisticated comprehension of every feature's influence on the result.

 Let us consider an example where a group of people (A, B, C, D, and E) work together on a project to make money (P) for their company. To guarantee that the team members receive an equal share of the company's profit according to their contributions to its achievement, it becomes necessary to determine each member's Shapley Value inside the group. Simply put, the difference between the profit made when a person 'A' participates in the team and when they don't is how one determines the person's Shapley Value. This computation provides information on the individual contribution of every team member, such as deciding Shapley values for features within a dataset. Fig. \ref{figS1} shown the process clearly. Fig. \ref{figS2} represents the calculation of different parameters for the Shapley values calculation. The SHAP value of member `A` is given by eqution 8, and the $\delta$ values are calculated as shown in Fig.\ref{figS2}.
 
 \begin{equation}
    \text{A}_{SHAP} = \frac{\delta_1 + \delta_2 + \delta_3 + \delta_4 + \delta_5 + \delta_6}{6}
\end{equation}
This discrepancy represents the "marginal contribution" of member 'A' to the group dynamics or ongoing coalition. The coalitions under discussion in this context are the different group configurations in which member 'A' participates. As such, the computation entails ascertaining member 'A's marginal contributions throughout every scenario for which a coalition could form. After averaging these marginal contributions, member 'A's Shapley Value is determined, and their impact on the group is thoroughly evaluated.

Fig. \ref{fig2} illustrates the complete workflow of our study, detailing the data journey from import to model evaluation. Initially, the raw data is imported to ensure accurate predictions, followed by necessary preprocessing and data cleaning. The processed data is then divided into training (70\%) and testing (30\%) sets. Subsequently, different models are constructed and trained using the training dataset. Hyperparameter tuning is performed to optimize model performance by iterating through various parameter values for optimal results.

After training the models, we use the test dataset to predict outcomes and evaluate their accuracy. We introduce an XAI method,namely SHAP, combined with GA and PCT based feature importance technique, to propose a novel feature evaluator algorithm AFE. AFE identifies key features and offers important explanations for the models. Since AFE is trained on testing data, it provides insights into the decision-making process of our best-performing model. Finally, the most suitable deployment model is selected based on comprehensive evaluation criteria and AFE's interpretability.
\subsection{Adaptive Feature Evaluator}
The Adaptive Feature Evaluator (AFE) is a novel algorithm designed to determine feature significance across various datasets. AFE consistently outperforms existing feature selection algorithms in comparative tests when applied to machine learning (ML) and deep learning (DL) models. Due to its global adaptability, AFE integrates seamlessly with any algorithm and dataset.

AFE combines three effective feature selection strategies: Explainable AI (XAI), Permutation \& Combination Techniques (PCT), and Genetic Algorithms (GA). It computes accuracy scores independently for these methods and synthesizes them, resulting in consistently higher accuracy. The robustness of AFE was assessed using six different algorithms, including Gradient Boosting, Multi-layer Perceptron Classifier, Gaussian Naive Bayes, Decision Tree, and Logistic Regression, and validated on three healthcare datasets: Covid-19, Heart Disease, and Lung Cancer. AFE demonstrated superior performance and flexibility in all cases compared to state-of-the-art methods.

\begin{figure*}[!t]
\centering
\includegraphics[width=6.0in]{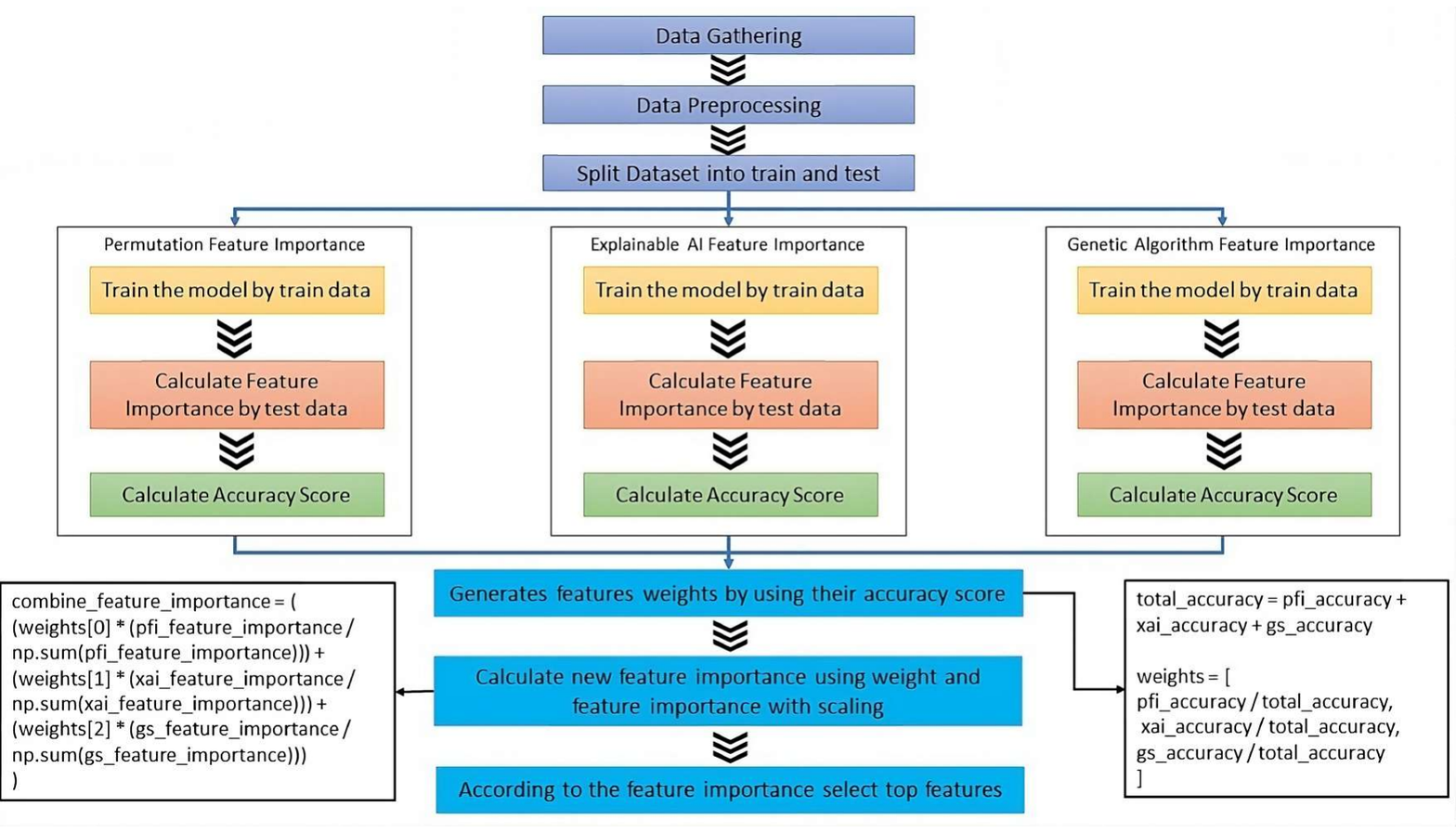}
\caption{AFE (Adaptive Feature Evaluator) algorithm flowchart}
\label{fig_AFE}
\end{figure*}

Through rigorous testing and validation, AFE establishes itself as a highly adaptive and accurate feature selection system, surpassing conventional techniques. Fig. \ref{fig_AFE} and Algorithm \ref{alg:AFE} clearly discuss the proposed AFE algorithm through flowchart and algorithm consecutively.
\begin{algorithm}
\caption{Proposed Adaptive Feature Evaluator Algorithm}
\label{alg:AFE}
\begin{algorithmic}[1]

    \STATE \textbf{Step 1: Input}
    \STATE Dataset $X$ (features), Target variable $y$
    
    \STATE \textbf{Step 2: Data Preparation}
    \STATE Split data into training set $(X_{\text{train}}, y_{\text{train}})$ and testing set $(X_{\text{test}}, y_{\text{test}})$
    
    \STATE \textbf{Step 3: Permutation \& Combination Technique (PCT)}
    \STATE Compute permutation importance on $X_{\text{test}}$
    \STATE Select features where importance $>$ median importance
    \STATE Train classifier using selected features
    \STATE Compute accuracy $PFI_{\text{accuracy}}$ on $X_{\text{test}}$
    
    \STATE \textbf{Step 4: SHAP Values}
    \STATE Compute SHAP values for $X_{\text{train}}$
    \STATE Select features where SHAP value $>$ median SHAP value
    \STATE Train classifier using selected features
    \STATE Compute accuracy $SHAP_{\text{accuracy}}$ on $X_{\text{test}}$
    
    \STATE \textbf{Step 5: Genetic Algorithm (GA) for Feature Selection}
    \STATE Apply Genetic Algorithm for feature selection on $X_{\text{train}}$
    \STATE Select features based on GA output
    \STATE Train classifier using selected features
    \STATE Compute accuracy $GA_{\text{accuracy}}$ on $X_{\text{test}}$
    
    \STATE \textbf{Step 6: Calculate Weights}
    \STATE Compute total accuracy: $Total_{\text{accuracy}} = PFI_{\text{accuracy}} + SHAP_{\text{accuracy}} + GA_{\text{accuracy}}$
    \STATE Compute individual weights:
    \STATE $Weight_{\text{PCT}} = \frac{PFI_{\text{accuracy}}}{Total_{\text{accuracy}}}$
    \STATE $Weight_{\text{SHAP}} = \frac{SHAP_{\text{accuracy}}}{Total_{\text{accuracy}}}$
    \STATE $Weight_{\text{GA}} = \frac{GA_{\text{accuracy}}}{Total_{\text{accuracy}}}$
    
    \STATE \textbf{Step 7: Combine Feature Importances}
    \STATE Normalize and combine feature importances from PCT, SHAP, and GA using the calculated weights:
    \STATE $Combined_{\text{importance}} = (Weight_{\text{PCT}} \times PCT_{\text{importance}} + Weight_{\text{SHAP}} \times SHAP_{\text{importance}} + Weight_{\text{GA}} \times GA_{\text{importance}})$
    
    \STATE \textbf{Step 8: Output}
    \STATE Sort features by $Combined_{\text{importance}}$ in descending order
    \STATE Display the sorted feature importance ranking

\end{algorithmic}
\end{algorithm}
\section{Experimental Results}
In this work, we propose and investigate the critical role of feature explainers using Explainable AI (XAI), in conjunction with the well-known Genetic Algorithm (GA) and Permutation Combination Technique (PCT), to enhance the performance and interpretability of machine learning algorithms. The evaluation is conducted across three distinct healthcare datasets—COVID-19, heart disease, and lung cancer, representing various medical conditions. This comprehensive assessment demonstrates the efficacy of the proposed Adaptive Feature Evaluator (AFE) across various healthcare domains.
\subsection{Simulation Setup}
The data visualization and manipulation experiment is in a Python 3.12 environment using packages like matplotlib, pandas, numpy, and seaborn. 

The implementation uses the scikit-learn (SKLearn) toolkit for predictive modeling, employing ten distinct algorithms to improve the accuracy of our predictions. We investigate many methods in the proposed studies because each algorithm is implemented by importing the required customized library to meet its particular needs.
We utilize the SHAP libraries and GA and PCT methods to provide our models with interpretability and insights. With the help of these libraries, we clarify the predictions made by our models, illuminating the underlying causes of the results and improving the transparency of our research. 
\subsection{Dataset Overview}
The proposed work use three different types of datasets for this investigation.
The Lung Cancer online repository from the University of California, Irvine \cite{misc_lung_cancer_62}. The entire dataset consists of one class attribute, 32 instances, and 57 characteristics. There are one label feature and fifteen input features in this dataset. Total 16 features and 309 data samples are available in the dataset.

The UCI Machine Learning Repository provided one of the datasets use in our study, which is centered on heart disease \cite{misc_heart_disease_45}. The output features in this dataset indicate several kinds of cardiac disorders. Total 12 features and 918 samples are available in the dataset.

One of the most extensive compilations of current COVID-19-related data is the Google Health COVID-19 Open Data Repository. Including information from over 20,000 sites globally, it offers a wide range of data formats to support researchers, policymakers, public health experts, and others in understanding and managing the virus. Total 11 features and 278848 samples are available in the dataset.

\subsection{Detailed Discussion}
This paper explores the role of feature importance in enhancing the performance and interpretability of machine learning algorithms across three diverse healthcare datasets—COVID-19, heart disease, and lung cancer. These datasets represent a broad spectrum of medical conditions, allowing for a comprehensive evaluation of the effectiveness of Explainable Artificial Intelligence (XAI) across different healthcare domains.

We employed six distinct classification algorithms: Logistic Regression (LR), Decision Tree (DT), Gaussian Naive Bayes (GNB), Random Forest (RF), Multi-layer Perceptron Classifier (MLP), Gradient Boosting (GB).
By leveraging this diverse set of algorithms, we aim to capture the variation in model performance across different healthcare datasets and conditions. Given that each algorithm offers unique strengths and weaknesses, this comprehensive approach is essential for thoroughly evaluating the models' performance and interpretability.

The proposed model ensemble consists of six algorithms, where the LR is with random\_state=0 (default parameters otherwise), Decision Tree with entropy criterion and random\_state=0, Gaussian Naive Bayes, Multinomial Naive Bayes, Random Forest are using default configurations. Additionally, we include an MLP Classified with a hidden layer size of 100, ReLU activation (f(x) = max(0, x)), Adam optimizer, and a learning rate of 0.001. Finally, the ensemble incorporates Gradient Boosting with log\_loss function, a learning rate of 0.1, and the friedman\_mse criterion.

The result matrices for COVID-19, heart disease, and lung cancer datasets are presented in Table \ref{Tab2}, where only data preprocessing was applied, without any feature importance techniques. These tables provide a baseline performance metric, enabling us to evaluate the impact of feature importance strategies in subsequent analyses. By comparing these initial results with those obtained after applying feature selection methods, we aim to demonstrate improvements in model accuracy, robustness, and interpretability. Performance is measured using accuracy and F1 score, with accuracy reflecting overall performance and F1 score balancing recall and precision as its harmonic mean.
\begin{figure*}[!t]
\centering
\includegraphics[width=6.2in]{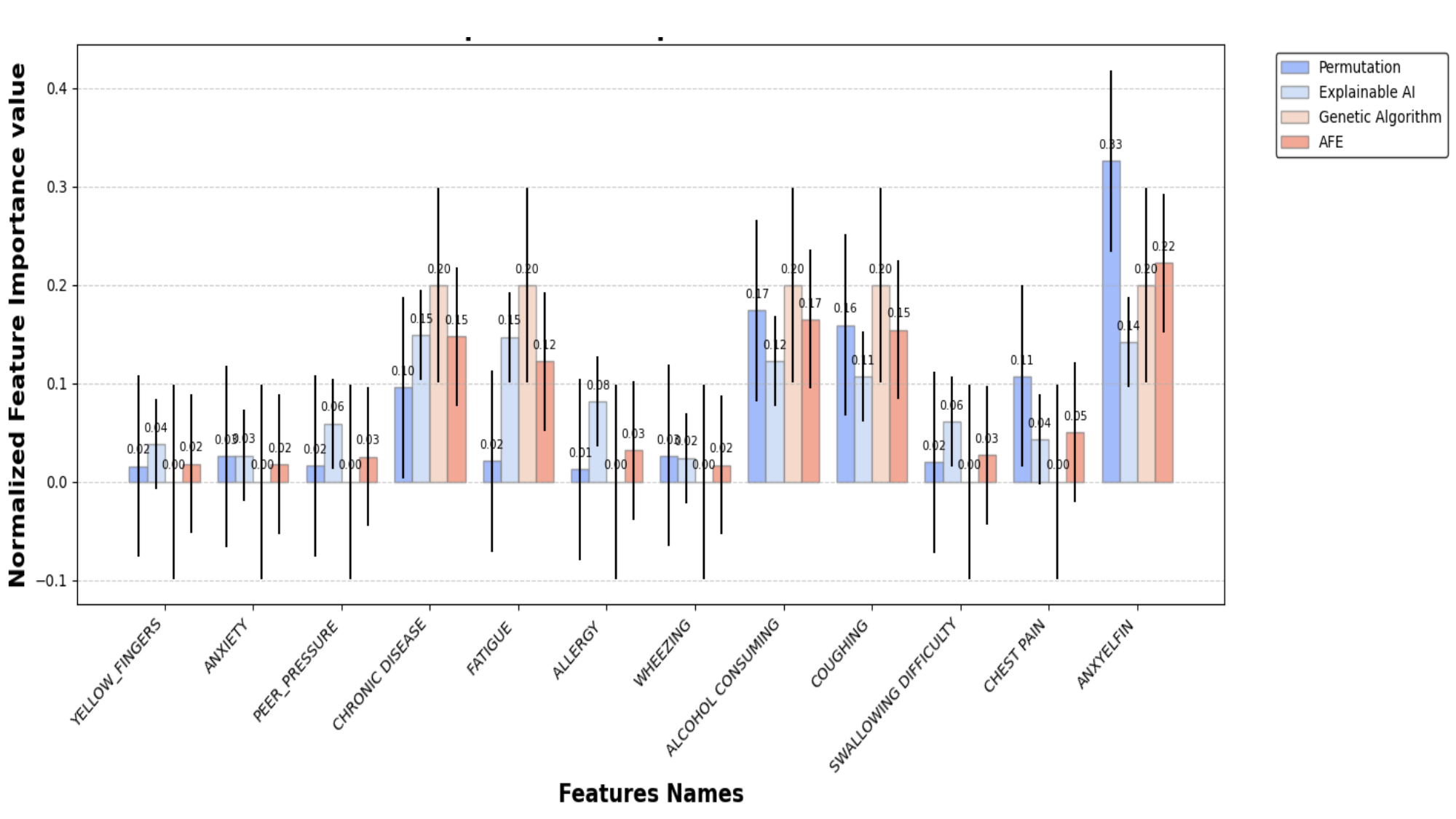}
\caption{AFE feature explanation for Lung Cancer Data}
\label{fig6}
\end{figure*}
\begin{figure*}[!t]
\centering
\includegraphics[width=6.2in]{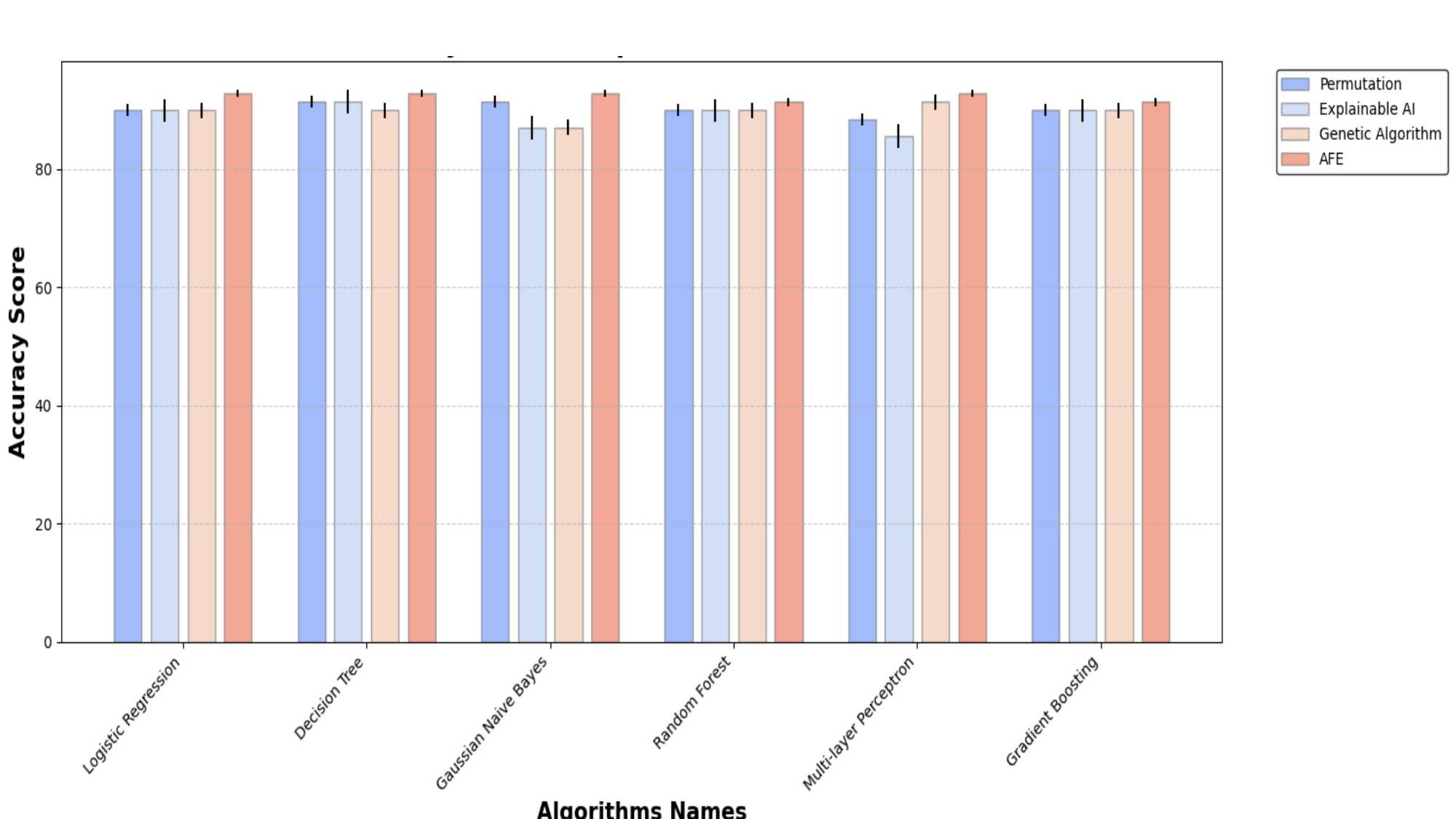}
\caption{Accuracy score comparision for Lung Cancer with AFE features}
\label{fig7}
\end{figure*}

\begin{table*}[htbp]
    \centering
    \caption{Performance prediction metrics without features importance consideration}
    \label{Tab2}
    \begin{tabular}{cccccccc}
    \hline \hline
  \multirow{2}{*}{SlNo} & \multirow{2}{*}{Algorithm} & \multicolumn{2}{c|}{Lung Cancer} & \multicolumn{2}{c|}{Heart Disease} & \multicolumn{2}{c}{Covid-19 Data} \\
    \cline{3-8}
    && Accuracy& F1 Score& Accuracy& F1 Score& Accuracy& F1 Score\\
    \hline
    1 & Logistic Regression & 89.855 &94.117  &83.478  &86.619 &92.363&92.526 \\[1ex]
    2 & Decision Tree &91.304  &94.915  &80.869  &83.823 &93.395&94.452\\[1ex]
   3 & Gaussian Naive Bayes &91.304  & 94.827 &83.478  &86.713 &90.452&91.352\\[1ex]
    4 & Random Forest & 91.304 & 94.915 & 83.043 & 85.920 &93.399&94.459\\[1ex]
    9 & Multilayer Perceptron&88.405  &93.333  & 84.782 &87.632 &92.396&94.452\\[1ex]
     10 & Gradient Boosting & 89.855 & 94.117 & 88.043 &86.021 &93.399&94.457\\[1ex]
      \hline
    \hline
    \end{tabular}    
   \end{table*}

\begin{table}[htbp]
    \centering
    \caption{Performance prediction metrics with features importance consideration for Lung Cancer}
    \label{Tab3}
    \begin{tabular}{cccccccccc}
    \hline \hline
  \multirow{2}{*}{Sl No} & \multirow{2}{*}{Algorithm} & \multicolumn{2}{c}{PCT} & \multicolumn{2}{c}{XAI} & \multicolumn{2}{c}{GA}& \multicolumn{2}{c}{AFE} \\
 \cline{3-10}
    && Accuracy& F1 Score& Accuracy& F1 Score& Accuracy& F1 Score& Accuracy& F1 Score\\
    \hline
    1 & LR & 89.955 &94.308  &89.855  &94.214 &89.855&94.308 &92.753&95.934\\[1ex]
    2 & DT &91.804  &95.915  &89.855  &94.017 &92.753&95.798&94.753&96.726\\[1ex]
   3 & GNB &86.956  & 92.307 &86.956  &92.307 &86.956&92.307&92.754&95.945\\[1ex]
    4 & RF & 89.855 & 94.117 & 92.753 & 95.867 &92.753&95.798&95.304&97.915\\[1ex]
    9 & MLP&89.855  &94.117  & 91.304 &95.081 &88.405&93.548&91.304&95.161\\[1ex]
     10 & GB & 89.855 & 94.117 & 89.855 &94.017 &92.753&95.798&92.314&94.916\\[1ex]
      \hline
    \hline
    \end{tabular}    
   \end{table}
\begin{figure*}[!t]
\centering
\includegraphics[width=6.2in]{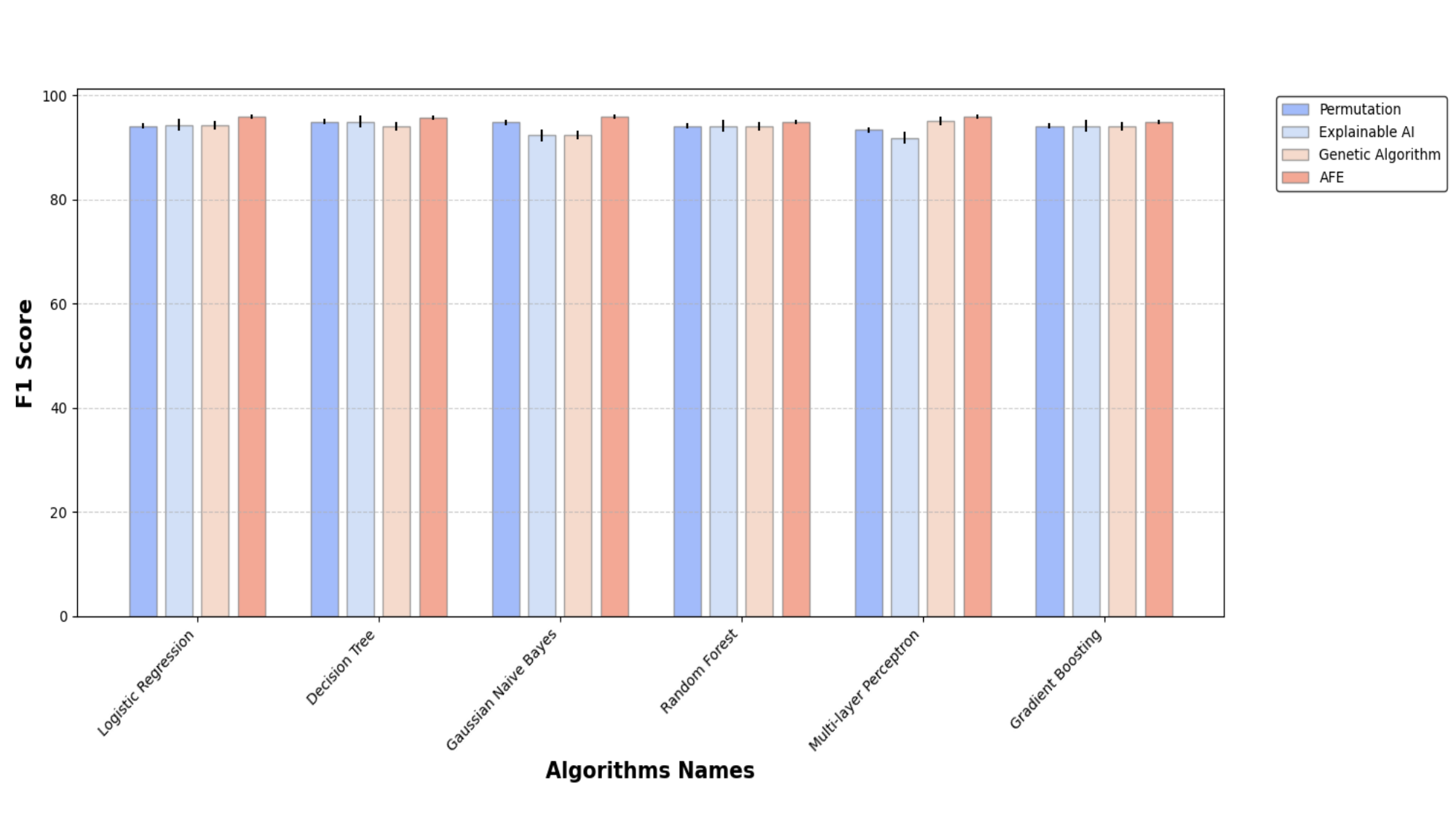}
\caption{F1 score comparision for Lung Cancer with AFE features}
\label{fig8}
\end{figure*}
\begin{figure*}[!t]
\centering
\includegraphics[width=6.2in]{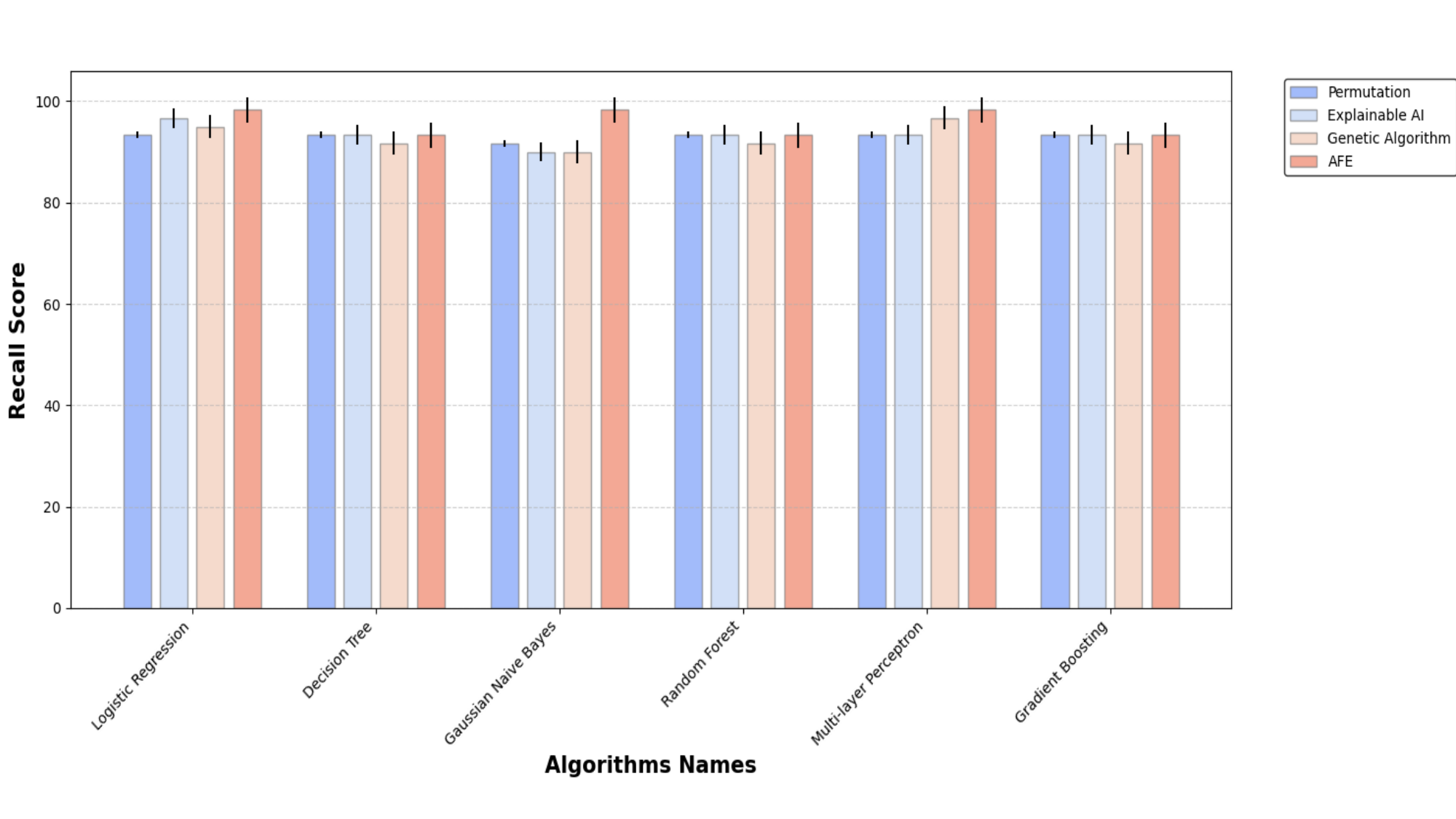}
\caption{Recall score comparision for Lung Cancer with AFE features}
\label{fig9}
\end{figure*}
\begin{figure*}[!t]
\centering
\includegraphics[width=6.2in]{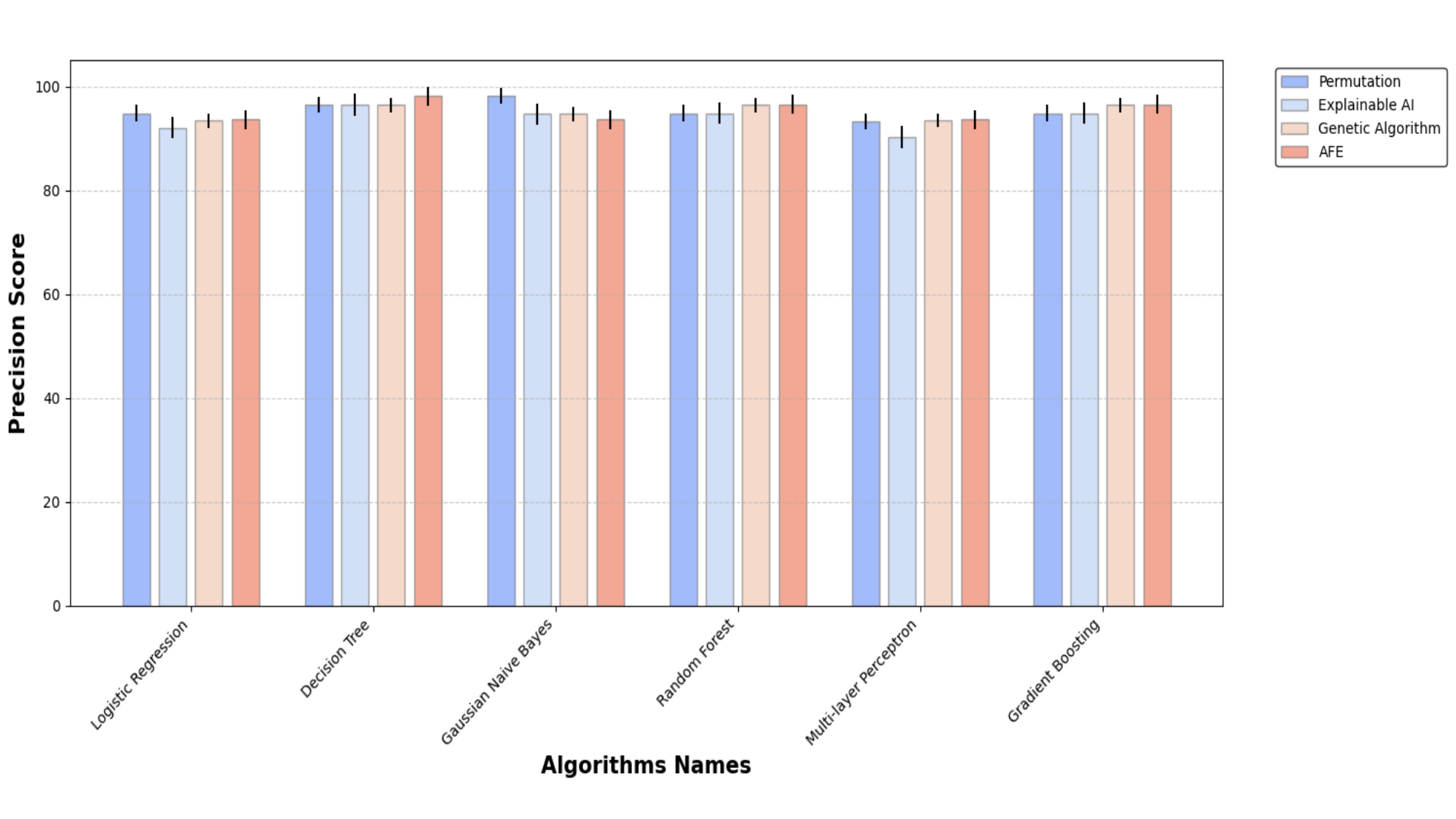}
\caption{Precision score comparision for Lung Cancer with AFE features}
\label{fig10}
\end{figure*}
\begin{table}[htbp]
    \centering
    \caption{Performance prediction metrics with AFE features importance consideration}
    \label{Tab4}
    \begin{tabular}{cccccc}
    \hline \hline
  \multirow{2}{*}{Sl No} & \multirow{2}{*}{Algorithm}  & \multicolumn{2}{c}{Heart Deasease} & \multicolumn{2}{c}{Covid-19 Data} \\
 \cline{3-6}
    && Accuracy& F1 Score& Accuracy& F1 Score\\
    \hline
    1 & Logistic Regression & 84.478 &86.330  &94.347  &96.234  \\[1ex]
    2 & Decision Tree &82.434  &83.985  &95.956  &97.586 \\[1ex]
   3 & Gaussian Naive Bayes &86.217  & 87.857 &92.782  &95.719 \\[1ex]
    4 & Random Forest & 89.521 & 91.967 & 97.652 & 99.421 \\[1ex]
    9 & Multilayer Perceptron&87.217  &88.943  & 98.517 &99.943 \\[1ex]
     10 & Gradient Boosting & 90.521 & 92.888 & 96.391 &98.975 \\[1ex]
      \hline
    \hline
    \end{tabular}    
   \end{table}
\begin{table}[htbp]
    \centering
    \caption{Features important of Lung Cancer Dataset}
    \label{Tab5}
    \begin{tabular}{ccc}
        \hline \hline
        Sl. No&Features Name & Feature weight\\[1ex]
       \hline
        1& ANXYELFIN&0.203482  \\[1ex]
        2&COUGHING & 0.135953 \\[1ex]
        3&CHRONIC DISEASE &  0.129164\\[1ex]
        4&FATIGUE & 0.103765 \\[1ex]
        5&ALCOHOL CONSUMING &0.099770  \\[1ex]
       6 &PEER\_PRESSURE & 0.072191 \\[1ex]
       7 &YELLOW\_FINGERS & 0.064810 \\[1ex]
        8&ANXIETY &0.064340  \\[1ex]
       9 & CHEST PAIN& 0.050378 \\[1ex]
       10 &ALLERGY & 0.031955\\[1ex]
        11 &SWALLOWING DIFFICULTY &0.027210  \\[1ex]
        12&WHEEZING &  0.016981\\[1ex]
       \hline \hline
    \end{tabular}
\end{table}
The accuracy metrics after applying Permutation Feature Importance, Explainable AI (XAI) techniques, and the Genetic Algorithm (GA) feature importance technique are presented in Table \ref{Tab3} using lung cancer data. While these methods improved the accuracy of specific models, they did not enhance performance across all algorithms. We introduced our proposed Adaptive Feature Evaluator (AFE) algorithm to address this limitation, consistently improving accuracy across all models and delivering more reliable and precise outcomes. By leveraging the strengths of multiple feature importance techniques, the AFE algorithm ensures robust performance and enhanced interpretability across diverse healthcare applications. Comparing the results obtained using the AFE technique, as shown in the final column of Table \ref{Tab3}, with the baseline metrics in Table \ref{Tab2}, it is evident that the AFE algorithm outperforms the other six models.

The Adaptive Feature Evaluator (AFE) was validated using two additional datasets: Heart disease and COVID-19. The accuracy metrics for these datasets are shown in Table \ref{Tab4}. In both cases, AFE consistently outperformed other feature significance techniques. This comprehensive evaluation demonstrates that AFE is a valuable tool for enhancing the performance of machine learning models in healthcare applications. It maintains robustness across diverse datasets and methods and significantly improves accuracy.

The feature selection process of the Adaptive Feature Evaluator (AFE) algorithm provides a relevance score for each feature, ranked in descending order, as shown in Table \ref{Tab5}. These feature importance scores are expressed as probabilities ranging from 0 to 1. The table ranks features by importance based on the Adaptive Feature Evaluator (AFE) algorithm. ANXYELFIN is the most significant feature (weight: 0.203), followed by COUGHING (0.136) and CHRONIC DISEASE (0.129). Features like FATIGUE (0.104) and ALCOHOL CONSUMING (0.100) are moderately important, while WHEEZING (0.017) is the least influential. This probabilistic representation offers an intuitive and straightforward understanding of each feature's relative importance, facilitating informed decision-making during model development. Figure \ref{fig6} shows the feature importance overview of lung cancer data. By highlighting the most critical features, the AFE algorithm enhances the interpretability and performance of machine learning models across various healthcare applications.
  



\begin{table}[h]
\centering
\caption{Performance comparison of proposed and previous research}
\begin{tabular}{cccc}
\hline\hline
Work & Dataset & Algorithm & Accuracy\\[1ex]
\hline
Ahmad et al.(2023)\cite{ahmed2023interpretable} & Lung Cancer  & Random Forest  & 97 \\
&Dataset&&\\[2ex]
Naseer et al.(2019)\cite{nasser2019lung}  & Lung Cancer  & ANN & 96.67 \\
&Dataset&&\\[2ex]
Dangare et al.(2012)\cite{dangare2012improved}  & UCI Heart  & KNN & 87.5 \\
&disease dataset&&\\[2ex]
Dwivedi et al.(2018)\cite{dwivedi2018performance}  & UCI Heart  & KNN &  80\\
&disease dataset&&\\[2ex]
Batista et al.(2020)\cite{batista2020covid}  & Covid-19  & SVM & 84.85 \\
&pandemic dataset&&\\[2ex]
Mahdy et al.(2020)\cite{mahdy2020automatic}  & Covid-19  & SVM & 95 \\
&pandemic dataset&&\\[2ex]
\hline
 & Lung Cancer  & Random Forest  & 95.5 \\
 &Dataset&&\\[2ex]
 Proposed Work&UCI Heart   & GB &  90.52\\
 &disease dataset&&\\[2ex]
 & Covid-19 pandemic  & MLP & 98.5 \\
 &dataset&&\\[2ex]
\hline\hline
\end{tabular}

\label{Tab8}
\end{table}


\subsection{Comparative Analysis with Existing Research}
The model proposed in the work achieves a prediction accuracy of 98.5\% on the Covid-19 Dataset using the MLP, outperforming the other models. Similarly, on the UCI Heart Disease Dataset, our model achieves an accuracy of 90.52\% using the GB algorithm, matching the performance of previous models. For the Lung Cancer Dataset, the proposed model acquires an accuracy of 95.5\% using the RF algorithm, demonstrating competitive performance compared to prior studies. Figure \ref{fig7} represents the effect of AFE features on others regarding accuracy. The comparison between the prediction accuracy of our proposed model and other existing models is shown in Table \ref{Tab8}. The table presents a performance comparison between the proposed work and previous research across different datasets and algorithms.

Figure \ref{fig8} illustrates the effect of AFE features on other models concerning F1 score. The comparison between the F1 score of our proposed model and other existing models is shown in the table. The table provides a performance comparison between the proposed work and previous research across various datasets and algorithms. Figure \ref{fig9} demonstrates the effect of AFE features on other models with respect to Recall. The comparison of the proposed model's Recall against other existing models is displayed in the table. This table compares the Recall performance of the proposed work and prior studies across different datasets and algorithms.

Figure \ref{fig10} shows the effect of AFE features on other models in terms of Precision. The table presents a comparison of the Precision of the proposed model versus other existing models, providing a performance comparison across various datasets and algorithms.

The proposed work demonstrates notable improvements in accuracy for each dataset and algorithm combination.
\section{Conclusion}
The work underscores the vital role of feature importance in enhancing the performance and interpretability of machine learning algorithms within the healthcare sector by evaluating various feature selection techniques across three distinct datasets: lung cancer, heart disease, and COVID-19. We identified the strengths and limitations of each method. The proposed Adaptive Feature Evaluator consistently surpasses traditional techniques, offering superior accuracy and robustness across diverse algorithms and datasets. This algorithm identifies key features with probabilistic importance and ensures their interpretability and relevance. The comprehensive evaluation validates the Adaptive Feature Evaluator as a transformative tool for advancing machine learning applications in healthcare, leading to more accurate, reliable, and actionable insights. As healthcare data grows in complexity, the Adaptive Feature Evaluator presents a promising solution for refining predictive models and enhancing clinical decision-making.

Prospective research will broaden the AFE algorithm's role in personalized medicine, particularly in oncology and complex healthcare domains, by integrating it with advanced machine learning models like deep learning and ensemble methods to enhance predictive accuracy and interpretability. Expanding AFE's application to diverse datasets, including real-time clinical and genomic data, will further assess its robustness. Incorporating explainability techniques such as SHAP and LIME alongside AFE could provide deeper insights into model predictions, which are essential for personalized treatment planning. Evaluating AFE with multi-modal data, including imaging, genetics, and electronic health records, may lead to more effective, patient-centric decision support systems, advancing precision medicine.
\bibliographystyle{unsrt}  

\section*{Author Biographies}

\begin{wrapfigure}{l}{28mm} 
\includegraphics[width=1in,height=1.25in,clip,keepaspectratio]{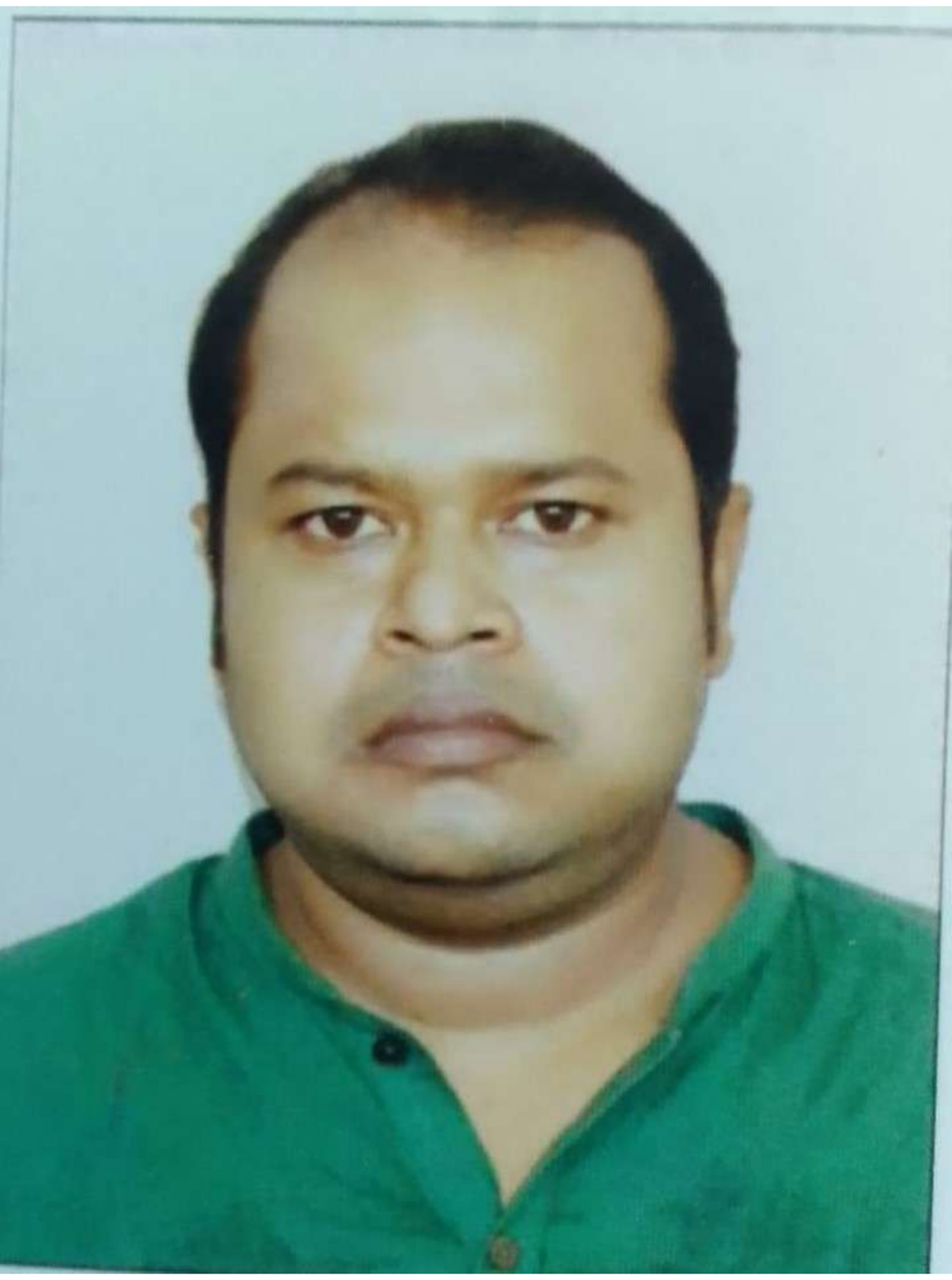}
\end{wrapfigure}\par
\textbf{Prasenjit Maji} (Member, IEEE) received the bachelor's degree in Computer Science \& Engineering from the West Bengal University of Technology, West Bengal, in 2007, the master's degree in Computer Science \& Engineering from the West Bengal University of Technology, West Bengal, in 2013, and is currently pursuing a Ph.D. from NIT, Durgapur. He worked as an Assistant Professor in the CSE Department, Bengal College of Engineering \& Technology, Durgapur, from 2013 to 2023 and is currently an assistant professor in the CSD Department, Dr. B. C. Roy Engineering College, Durgapur. His research interests include Image Processing and Machine Learning. His Google Scholar h-index is 10, and i-index is 10 with 275 citations.

\vspace{1cm}
\begin{wrapfigure}{L}{0.22\textwidth}
\vspace{-2cm}
\includegraphics[width=1in,height=1.4in,clip,keepaspectratio]{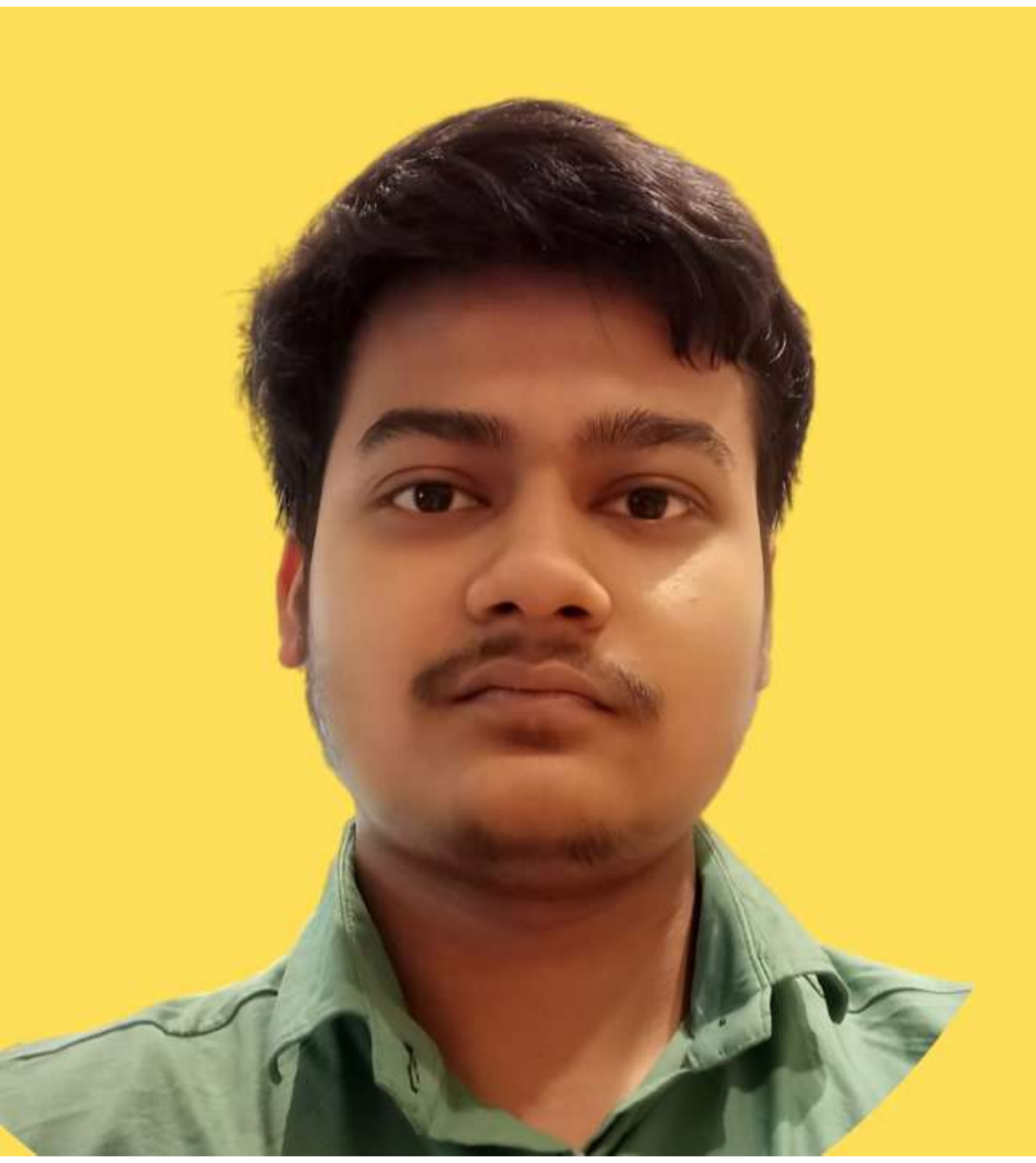}
\end{wrapfigure}\par
\textbf{Amit Kumar Mondal} is an undergraduate student pursuing a Bachelor of Technology in the Computer Science and Engineering Department of Bengal College of Engineering and Technology, Durgapur, India. His research interests include the development of robust machine learning and deep learning techniques for smart healthcare systems.

\begin{wrapfigure}{l}{28mm} 
\includegraphics[width=1in,height=1.48in,clip,keepaspectratio]{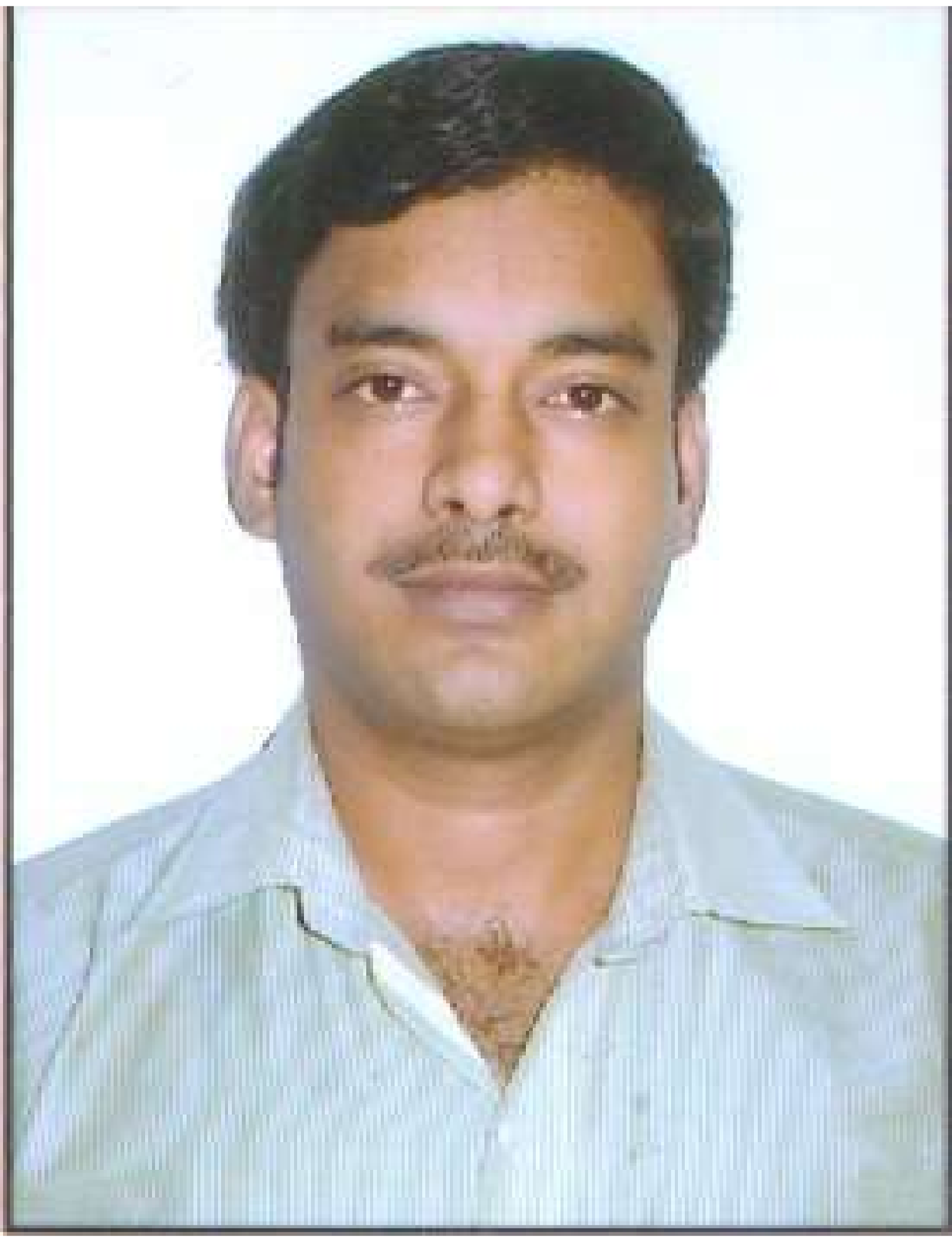}
\end{wrapfigure}\par
\textbf{Hemanta Kumar Mondal}(Senior Member, IEEE) received the bachelor's degree in electronics and communication engineering from the West Bengal University of Technology, West Bengal, in 2007, the master's degree in VLSI Design from Guru Gobind Singh Indraprastha University, Delhi, in 2010, and the Ph.D. degree in Electronics and Communication Engineering from IIIT, Delhi, in 2017. He was associated with the National Centre for Scientific Research (CNRS) Lab-STICC, UBS University in France as a Postdoctoral Researcher in 2017. He is currently an assistant professor at the National Institute of Technology, Durgapur, West Bengal. He has authored 45 research articles. His Google Scholar h-index is 15, and i-index is 20 with 622 citations.
\clearpage
\vspace{-1cm}
\begin{wrapfigure}{l}{25mm} 
\includegraphics[width=1in,height=1.28in,clip,keepaspectratio]{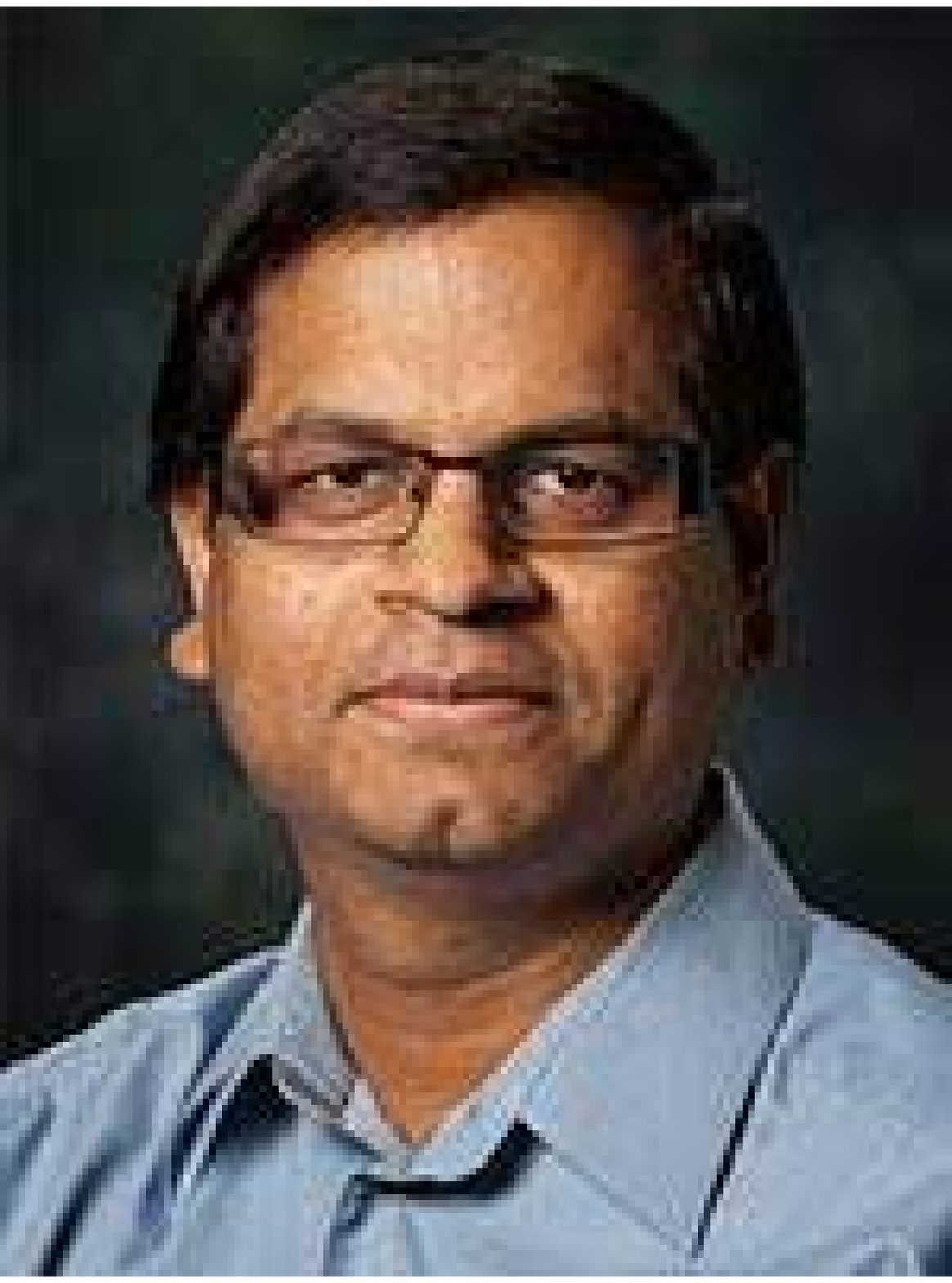}
\end{wrapfigure}\par
\textbf{Saraju P. Mohanty}(Senior Member, IEEE) received the bachelor’s degree (Honors) in electrical engineering from the Orissa University of Agriculture and Technology, Bhubaneswar, in 1995, the master’s degree in Systems Science and Automation from the Indian Institute of Science, Bengaluru, in 1999, and the Ph.D. degree in Computer Science and Engineering from the University of South Florida, Tampa, in 2003. He is a Professor with the University of North Texas. His research is in ``Smart Electronic Systems’’ which has been funded by National Science Foundations (NSF), Semiconductor Research Corporation (SRC), U.S. Air Force, IUSSTF, and Mission Innovation. He has authored 550 research articles, 5 books, and 10 granted and pending patents. His Google Scholar h-index is 58 and i10-index is 269 with 15,000 citations. He is regarded as a visionary researcher on Smart Cities technology in which his research deals with security and energy aware, and AI/ML-integrated smart components. He introduced the Secure Digital Camera (SDC) in 2004 with built-in security features designed using Hardware Assisted Security (HAS) or Security by Design (SbD) principle. He is widely credited as the designer for the first digital watermarking chip in 2004 and first the low-power digital watermarking chip in 2006. He is a recipient of 19 best paper awards, Fulbright Specialist Award in 2021, IEEE Consumer Electronics Society Outstanding Service Award in 2020, the IEEE-CS-TCVLSI Distinguished Leadership Award in 2018, and the PROSE Award for Best Textbook in Physical Sciences and Mathematics category in 2016. He has delivered 30 keynotes and served on 15 panels at various International Conferences. He has been serving on the editorial board of several peer-reviewed international transactions/journals, including IEEE Transactions on Big Data (TBD), IEEE Transactions on Computer-Aided Design of Integrated Circuits and Systems (TCAD), IEEE Transactions on Consumer Electronics (TCE), and ACM Journal on Emerging Technologies in Computing Systems (JETC). He has been the Editor-in-Chief (EiC) of the IEEE Consumer Electronics Magazine (MCE) during 2016-2021. He served as the Chair of Technical Committee on Very Large Scale Integration (TCVLSI), IEEE Computer Society (IEEE-CS) during 2014-2018 and on the Board of Governors of the IEEE Consumer Electronics Society during 2019-2021. He serves on the steering, organizing, and program committees of several international conferences. He is the steering committee chair/vice-chair for the IEEE International Symposium on Smart Electronic Systems (IEEE-iSES), the IEEE-CS Symposium on VLSI (ISVLSI), and the OITS International Conference on Information Technology (OCIT). He has supervised 3 post-doctoral researchers, 17 Ph.D. dissertations, 28 M.S. theses, and 28 undergraduate projects.

\end{document}